\documentclass[usenames,dvipsnames]{article} 
\usepackage{iclr2024_conference,times}

\usepackage{graphicx} 
\usepackage{booktabs}
\usepackage{colortbl}
\usepackage{multirow}
\usepackage{longtable}
\usepackage{lscape}
\usepackage{rotating}

\PassOptionsToPackage{usenames,dvipsnames}{xcolor}
\usepackage{xcolor}

\usepackage{scalefnt,letltxmacro}
\LetLtxMacro{\oldtextsc}{\textsc}
\renewcommand{\textsc}[1]{\oldtextsc{\scalefont{1.10}#1}}
\usepackage[acronym,nowarn]{glossaries}
\glsdisablehyper
\makeglossaries
\usepackage{xspace}
\usepackage{float}
\usepackage{physics}
\usepackage[normalem]{ulem}

\usepackage{enumitem}

\usepackage{amssymb}
\usepackage{mathtools}
\usepackage{amsfonts}
\usepackage{amsmath}
\usepackage{amsthm,thmtools}
\usepackage{booktabs}
\usepackage[]{microtype}

\usepackage[linesnumbered,ruled,vlined]{algorithm2e}

\SetCommentSty{mycommfont}
\SetKwInput{KwInput}{Input}
\SetKwInput{KwOutput}{Output}

\usepackage{hyperref}
\hypersetup{
  colorlinks   = true, 
  urlcolor     = blue, 
  linkcolor    = NavyBlue, 
  citecolor   = Mahogany 
}

\setcitestyle{square,semicolon}

\usepackage[capitalize,nameinlink,]{cleveref} 
\crefname{section}{\S}{\S\S}
\Crefname{section}{\S}{\S\S}
\Crefname{equation}{Eq.}{Eqs.}

\makeatletter
\@ifundefined{c@rownum}{%
	\let\c@rownum\rownum
}{}
\@ifundefined{therownum}{%
	\def\therownum{\@arabic\rownum}%
}{}
\makeatother

\makeatletter
\newcommand*{\addFileDependency}[1]{
	\typeout{(#1)}
	\@addtofilelist{#1}
	\IfFileExists{#1}{}{\typeout{No file #1.}}
}
\makeatother

\usepackage[font=footnotesize,labelfont=bf,tableposition=top]{caption}
\usepackage{tikz}
\usepackage{graphicx}
\usepackage[]{subcaption}   

\usepackage{pgfplots}
\pgfplotsset{compat=1.6}
\usepgfplotslibrary{groupplots}
\tikzstyle{every picture}+=[font=\sffamily]
\tikzstyle{optimized} = [circle,fill=white,draw=black, dashed,inner sep=1pt, minimum size=20pt, font=\fontsize{10}{10}\selectfont, node distance=1]
\pgfkeys{/pgf/number format/.cd,1000 sep={}}
\pgfplotsset{
	tick label style = {font=\sffamily},
	every axis label/.append style={font=\sffamily},
	typeset ticklabels with strut,
}
\pgfplotsset{every axis/.append style={
			every x tick label/.append style={font=\fontsize{6pt}{6pt}\sffamily, yshift=.5ex,},
			every y tick label/.append style={font=\fontsize{6pt}{6pt}\sffamily, xshift=.5ex},
			every y label/.append style={xshift=10ex, font=\sffamily},
			every x label/.append style={yshift=3ex, font=\sffamily},
			every title/.append style={font=\sffamily}
		},
}
\pgfplotsset{
	xticklabel={$\mathsf{\pgfmathprintnumber{\tick}}$},
	yticklabel={$\mathsf{\pgfmathprintnumber{\tick}}$},
}
\pgfplotsset{every axis title/.append style={yshift=-1ex}}
\newlength\figureheight
\newlength\figurewidth
\usepgfplotslibrary{external}
\tikzexternaldisable

\usepackage[colorinlistoftodos,
	textsize=scriptsize,
	linecolor=red!30,
	bordercolor=red!30,
	backgroundcolor=red!10]{todonotes}
\renewcommand{\todo}[2][]{\tikzexternaldisable\@todo[#1]{#2}\tikzexternalenable}

\usepackage[most]{tcolorbox}
{\begin{tcolorbox}[toprule=2mm,left=4pt,right=4pt,top=0pt,bottom=1pt,boxsep=2pt, before skip = 2ex, after skip =2ex]}{\end{tcolorbox}}
\tcbset{boxsep=4pt,left=2pt,right=2pt,top=-0pt,bottom=0pt}

\usepackage{url}

\usepackage{subcaption}

\usepackage{tikz}
\usepackage{graphicx}
\usepackage{amsmath}
\usepackage{amssymb}
\usepackage{bm}
\usepackage{amsthm} 
\usepackage{nicefrac}

\usepackage{adjustbox}
\usepackage{multirow}
\usepackage{mathtools}

\usepackage{xspace}

\usepackage{multirow}
\usepackage{multicol}

\newacronym{MC}{\textsc{mc}}{Monte Carlo}
\newacronym{SDE}{\textsc{sde}}{Stochastic Differential Equation}

\newacronym{KL}{\textsc{kl}}{Kullback-Leibler}

\newacronym{MI}{\textsc{mi}}{Mutual Information}
\newacronym{MINDE}{\textsc{minde}}{Mutual Information Neural Diffusion Estimation}

\newcommand{\g}{\,|\,}

\DeclareRobustCommand{\KL}[2]{\ensuremath{\textsc{kl}\left[#1\;\|\;#2\right]}}
\DeclarePairedDelimiterX{\infdivx}[2]{[}{]}{%
#1\;\delimsize\|\;#2%
}

\DeclareRobustCommand{\Gauss}[1]{\ensuremath{\mathcal{N}_{#1}}}

\newcommand{\E}{\mathbb{E}}

\newcommand{\defeq}{\stackrel{\text{\tiny def}}{=}}

\DeclareRobustCommand{\Gauss}[1]{\ensuremath{\bar{\gamma}_{#1}}}
\DeclareRobustCommand{\GaussM}[1]{\ensuremath{\gamma_{#1}}}

\DeclareMathAlphabet{\pazocal}{OMS}{zplm}{m}{n}

\usepackage[english]{babel}

\title{MINDE: Mutual Information Neural Diffusion Estimation}

\author{
Giulio Franzese$^{1,\star}$,
Mustapha Bounoua$^{1,2}$, Pietro Michiardi$^{1}$\\
{$^{1}$EURECOM,$^{2}$Ampere Software Technology},\quad
$^{\star}$\texttt{giulio.franzese@eurecom.fr}
}

\iclrfinalcopy

\begin{document}
\maketitle

\begin{abstract}

In this work we present a new method for the estimation of \gls{MI} between random variables. Our approach is based on an original interpretation of the Girsanov theorem, which allows us to use score-based diffusion models to estimate the \gls{KL} divergence between two densities as a difference between their score functions. As a by-product, our method also enables the estimation of the entropy of random variables. 
Armed with such building blocks, we present a general recipe to measure \gls{MI}, which unfolds in two directions: one uses conditional diffusion process, whereas the other uses joint diffusion processes that allow  simultaneous modelling of two random variables. 
Our results, which derive from a thorough experimental protocol over all the variants of our approach, indicate that our method is more accurate than the main alternatives from the literature, especially for challenging distributions. Furthermore, our methods pass \gls{MI} self-consistency tests, including data processing and additivity under independence, which instead are a pain-point of existing methods. 
\href{https://github.com/MustaphaBounoua/minde}{Code available.}

\end{abstract}

\section{Introduction}
Mutual Information (\gls{MI}) is a central measure to study the non-linear dependence between random variables \citep{shannon1948mathematical, mackay2003information}, and has been extensively used in machine learning for representation learning \citep{bell1995information, stratos2018mutual, belghazi2018mine, oord2019representation, hjelm2019learning}, and for both training \citep{alemi2019deep, chen2016infogan, zhao2018information} and evaluating generative models \citep{alemi2019gilbo, huang2020evaluating}. 

For many problems of interest, precise computation of \gls{MI} is not an easy task \citep{mcallester2020formal, paninski2003estimation}, and a wide range of techniques for \gls{MI} estimation have flourished. As the application of existing parametric and non-parametric methods \citep{pizer1987adaptive, moon1995estimation, kraskov2004estimating, gao2015efficient} to realistic, high-dimensional data is extremely challenging, if not unfeasible, recent research has focused on variational approaches \citep{barber2004algorithm, nguyen2007neurips, nowozin2016neurips, poole2019variational, wunder2021reverse, letizia2023variational, federici2023effectiveness} and neural estimators \citep{papamakarios2017masked, belghazi2018mine, oord2019representation, song2019understanding, rhodes2020telescoping, letizia2022copula, brekelmans2023improving} for \gls{MI} estimation. 
In particular, the work by \citet{song2019understanding} and \citet{federici2023effectiveness} classify recent \gls{MI} estimation methods into discriminative and generative approaches. The first class directly learns to estimate the ratio between joint and marginal densities, whereas the second estimates and approximates them separately. 

In this work, we explore the problem of estimating \gls{MI} using generative approaches, but with an original twist. In \Cref{sec:diffusion} we review diffusion processes \citep{song2021a} and in \Cref{sec:kl_deriv} we explain how, thanks to the Girsanov Theorem \citep{oksendal2003stochastic}, we can leverage score functions to compute the \gls{KL} divergence between two distributions. This also enables the computation of the entropy of a random variable. In \Cref{sec:mutualinfo} we present our general recipe for computing the \gls{MI} between two arbitrary distributions, which we develop according to two modeling approaches, i.e., conditional and joint diffusion processes. The conditional approach is simple and capitalizes on standard diffusion models, but it is inherently more rigid, as it requires one distribution to be selected as the conditioning signal. Joint diffusion processes, on the other hand, are more flexible, but require an extension of traditional diffusion models, which deal with dynamics that allow data distributions to evolve according to multiple arrows of time.

Recent work by \citet{czyz2023beyond} argue that \gls{MI} estimators are mostly evaluated assuming simple, multivariate normal distributions for which \gls{MI} is analytically tractable, and propose a novel benchmark that introduces several challenges for estimators, such as sparsity of interactions, long-tailed distributions, invariance, and high mutual information. Furthermore, \citet{song2019understanding} introduce measures of self-consistency (additivity under independence and the data processing inequality) for \gls{MI} estimators, to discern the properties of various approaches. In \Cref{sec:experiments} we evaluate several variants of our method, which we call \gls{MINDE}, according to such challenging benchmarks: our results show that \gls{MINDE} outperforms the competitors on a majority of tasks, especially those involving challenging data distributions. Moreover, \gls{MINDE} passes all self-consistency tests, a property that has remained elusive so far, for existing neural \gls{MI} estimators.

\section{Diffusion Processes and Score Functions}\label{sec:diffusion}
We now revisit the theoretical background on diffusion processes, which is instrumental for the derivation of the methodologies proposed in this work. 
Consider the real space $\mathbb{R}^N$ and its associated Borel $\sigma-$algebra, defining the measurable space $\left(\mathbb{R}^N,\mathcal{B}(\mathbb{R}^N)\right)$. In this work, we consider Ito processes in $\mathbb{R}^N$ with duration $T<\infty$. Let $\Omega=D\left([0,T]\times \mathbb{R}^N\right)$, be the space of all $N-$dimensional continuous functions in the interval $[0,T]$, and the filtration $\mathcal{F}$ induced by the canonical process $X_t(\omega)=\omega_t,\omega \in \Omega$. As starting point, we consider an Ito process:
\begin{equation}\label{eq:diffsde_deter}
    \begin{cases}
    \dd X_t=f_tX_t\dd t+ g_t\dd W_t,\\
    X_0=x
    \end{cases}
\end{equation}
with given continuous functions $f_t\leq 0, g_t>0$ and an arbitrary (deterministic) initial condition $x\in \mathbb{R}^N$. 
Equivalently, we can say that initial conditions are drawn from the Dirac measure $\delta_x$. This choice completely determines the path measure $\mathbb{P}^{\delta_x}$ of the corresponding probability space $\left(\Omega,\mathcal{F},\mathbb{P}^{\delta_x}\right)$. Starting from $\mathbb{P}^{\delta_x}$ we construct a new path measure $\mathbb{P}^{\mu}$ by considering the product between $\mathbb{P}^{\delta_x}$ and measure $\mu$ in $\mathbb{R}^N$:
\begin{equation}\label{prodmeas}
\mathbb{P}^{\mu}=\int\limits_{\mathbb{R}^N}\mathbb{P}^{\delta_x} \dd \mu(x).
\end{equation}
Conversely, the original measure $\mathbb{P}^{\delta_x}$ can be recovered from $\mathbb{P}^{\mu}$ by conditioning the latter on the particular initial value $x$, i.e., the \textit{projection} $\mathbb{P}^{\delta_x}=\mathbb{P}^{\mu}\#_x$. The new measure $\mathbb{P}^{\mu}$ can be represented by the following \gls{SDE}:
\begin{equation}\label{eq:diffsdes}
    \begin{cases}
    \dd X_t=f_tX_t\dd t+ g_t\dd W_t,\\
    X_0\sim \mu
    \end{cases}
\end{equation}
associated to the corresponding probability spaces $\left(\Omega,\mathcal{F},\mathbb{P}^{\mu}\right)$. We define $\nu^{\mu}_t$ as the pushforward of the complete path measure onto time instant $t\in[0,T]$, where by definition $\nu^{\mu}_0=\mu$.

It is instrumental for the scope of this work to study how the path measures and the \glspl{SDE} representations change under \textit{time reversal}.
Let $\hat{X}_t\defeq \omega_{T-t}$ be the time-reversed canonical process. 
If the canonical process $X_t$ is represented as in \Cref{eq:diffsdes} under the path measure $\mathbb{P}^{\mu}$, then the time reversed process $\hat{X}_t$ has \gls{SDE} representation \citep{anderson1982reverse}:
\begin{equation}\label{eq:diffsdes2}
    \begin{cases}
    \dd \hat X_t=-f_{T-t}\hat X_t+g^2_{T-t}s^{\mu}_{T-t}(\hat X_t)\dd t+ g_{T-t}\dd \hat W_t,\\
    \hat X_0\sim \nu^{\mu}_T
    \end{cases}
\end{equation}
with corresponding path-reversed measure $\hat{\mathbb{P}}^{\mu}$, on the probability spaces with time-reversed filtration. 

Next, we define the \textbf{score function} of the densities associated to the forward processes. In particular, $s^{\mu}_t(x)\defeq\nabla\log \left(\bar{\nu}^{\mu}_{t}(x)\right)$, where $\bar{\nu}^{\mu}_t(x)$ is the density associated to the measures $\nu^{\mu}_t(x)$, computed with respect to the Lebesgue measure,
 $\dd{\nu}^{\mu}_t(x)=\bar{\nu}^{\mu}_t(x)\dd x$.
In general we cannot assume exact knowledge of such \textit{true} score function.
Then, in practice, instead of the score function $s^{\mu}_t(x)$, we use \textit{parametric} ($\theta$) approximations thereof, $\tilde{s}^{\mu}_t(x)$, which we call the score network. Training the score network can be done by minimizing the following loss \citep{song2021a,huang2021variational,kingma2021}:
\begin{equation}\label{eq:loss}
    \mathcal{L}(\theta)=\E_{   {\mathbb{P}}^{\mu}}\left[\int\limits_0^T\frac{g^2_{t}}{2}\norm{\tilde{s}^{\mu}_t(X_t)-\nabla\log \left(\bar{\nu}^{\delta_{X_0}}_{t}(X_t)\right)}^2
    \dd t\right],
\end{equation}
where $\nu_t^{\delta_{X_0}}$ stands for the measure of the processes at time $t$, conditioned on some initial value $X_0$.

\section{KL divergence as difference of Score Functions}\label{sec:kl_deriv}
The \gls{MI} between two random variables can be computed according to several equivalent expressions, which rely on the \gls{KL} divergence between measures and/or entropy of measures. 
We then proceed to describe i) how to derive \gls{KL} divergence between measures as the expected difference of score functions, ii) how to estimate such divergences given parametric approximation of the scores (and the corresponding estimation errors) and iii) how to cast the proposed methodology to the particular case of entropy estimation. In summary, this Section introduces the basic building blocks that we use in \Cref{sec:mutualinfo} to define our \gls{MI} estimators. 

We consider the \gls{KL} divergence between two generic measures $\mu^A$ and $\mu^B$ in $\mathbb{R}^N$, i.e. $\KL{\mu^A}{\mu^B}$, which is equal to $\int_{\mathbb{R}^N} \dd\mu^A\log(\frac{\dd\mu^A}{\dd\mu^B})$, if the Radon-Nikodym derivative $\frac{\dd\mu^A}{\dd\mu^B}$ exists (absolute continuity is satisfied), and $+\infty$ otherwise. Since our state space is $\mathbb{R}^N$, the following \textit{disintegration} properties are valid \citep{leonard2014some}:
\begin{equation}\label{eq:disintegration}
    \frac{\dd \mathbb{P}^{\mu^{A}}}{\dd \mathbb{P}^{\mu^{B}}} (\omega)=\frac{\dd \left(\mathbb{P}^{\mu^A}\#_{\omega_{0}}\right)}{\dd \left(\mathbb{P}^{\mu^B}\#_{\omega_{0}}\right)} (\omega)\frac{\dd \mu^A(\omega_{0})}{\dd \mu^B(\omega_{0})}=\frac{\dd \mu^A(\omega_{0})}{\dd \mu^B(\omega_{0})},
    \frac{\dd\hat{\mathbb{P}}^{\mu^{A}}}{ \dd \hat{\mathbb{P}}^{\mu^B}} (\omega)=
    \frac{\dd\left(\hat{\mathbb{P}}^{\mu^A}\#_{\omega_{T}} \right) }{\dd\left(\hat{\mathbb{P}}^{\mu^B}\#_{\omega_{T}} \right)} (\omega)
    \frac{\dd \nu^{\mu^A}_T(\omega_{T})}{\dd \nu^{\mu^B}_T(\omega_{T})},
\end{equation}

where we implicitly introduced the product representation  $\hat{\mathbb{P}}^{\mu^{A}}=\int\limits_{\mathbb{R}^N}\hat{\mathbb{P}}_x \dd \nu^{\mu^A}_T(x)$, similarly to \Cref{prodmeas}. Thanks to such disintegration theorems, we can write the \gls{KL} divergence between the overall path measures $\mathbb{P}^{\mu^{A}}$ and $\mathbb{P}^{\mu^{B}}$ of two diffusion processes associated to the measures $\mu^A$ and $\mu^B$ as
\begin{flalign}
       &\KL{\mathbb{P}^{\mu^{A}}}{\mathbb{P}^{\mu^{B}}}=\E_{\mathbb{P}^{\mu^{A}}}\left[\log\frac{\dd\mathbb{P}^{\mu^{A}}}{\dd\mathbb{P}^{\mu^{B}}}\right]=\E_{\mathbb{P}^{\mu^{A}}}\left[\log\frac{\dd\mu^{A}}{\dd\mu^{B}}\right]=\KL{\mu^A}{\mu^B},\label{eq:kl1} 
\end{flalign}
where the second equality holds because, as observed on the left of \Cref{eq:disintegration}, when conditioned on the same initial value, the path measures of the two forward processes coincide. 

Now, since the \gls{KL} divergence between the path measures is invariant to time reversal, i.e., $\KL{\mathbb{P}^{\mu^{A}}}{\mathbb{P}^{\mu^{B}}}=  \KL{\hat{\mathbb{P}}^{\mu^{A}}}{\hat{\mathbb{P}}^{\mu^{B}}}$, using similar disintegration arguments, it holds that:

\begin{equation}\label{eq:kl2}
 \KL{\hat{\mathbb{P}}^{\mu^{A}}}{\hat{\mathbb{P}}^{\mu^{B}}}=  \E_{   \hat{\mathbb{P}}^{\mu^{A}}}\left[\log\frac{\dd\left( \hat{\mathbb{P}}^{\mu^{A}}\#_{\omega_{T}}\right) }{\dd\left(\hat{\mathbb{P}}^{\mu^{B}}\#_{\omega_{T}}\right)}\right]
    +\E_{   \hat{\mathbb{P}}^{\mu^{A}}}\left[\log\frac{\dd   \nu^{\mu^A}_T}{\dd\nu^{\mu^B}_T}\right].
\end{equation}

The first term on the r.h.s of \Cref{eq:kl2} can be computed using the Girsanov theorem \citep{oksendal2003stochastic} as
\begin{flalign}
  &\E_{   \hat{\mathbb{P}}^{\mu^{A}}}\left[\int\limits_0^T\frac{1}{2g^2_{t}}\norm{g^2_{t}\left(s^{\mu^A}_t(\hat X_t)-s^{\mu^B}_t(\hat X_t)\right) }^2\dd t\right]=\E_{   {\mathbb{P}}^{\mu^A}}\left[\int\limits_0^T\frac{g^2_{t}}{2}\norm{s^{\mu^A}_t(X_t)-s^{\mu^B}_t(X_t) }^2\dd t\right].  \label{girsanov}
\end{flalign}

The second term on the r.h.s of \Cref{eq:kl2}, equals $\KL{\nu^{\mu^A}_T}{\nu^{\mu^B}_T}$: this is a vanishing term with $T$, i.e. $\lim_{T\rightarrow\infty}\KL{\nu^{\mu^A}_T}{\nu^{\mu^B}_T}=0$. To ground this claim, 
we borrow the results by \citet{collet2008logarithmic}, which hold for several forward diffusion \glspl{SDE} of interest, such as the Variance Preserving (VP), 
or Variance Exploding (VE) \glspl{SDE} \cite{song2021a}.
In summary, it is necessary to adapt the classical Bakry-Émery condition of diffusion semigroup to the non homogeneous case, and exploit the contraction properties of diffusion on the KL divergences. 

Combining the different results, we have that: 
\begin{equation}\label{kl_formula}
    \KL{\mu^A}{\mu^B}=\E_{   {\mathbb{P}}^{\mu^A}}\left[\int\limits_0^T\frac{g^2_{t}}{2}\norm{s^{\mu^A}_t(X_t)-s^{\mu^B}_t(X_t) }^2\dd t\right]+\KL{\nu^{\mu^A}_T}{\nu^{\mu^B}_T}
\end{equation}
which constitutes the basic equality over which we construct our estimators, described in \Cref{sec:estimators}.

We conclude by commenting on the possibility of computing divergences in a \textit{latent} space. Indeed, in many natural cases, the density $\mu^A$ is supported on a lower dimensional manifold $\mathcal{M}\subset \mathbb{R}^N$~\citep{loaiza2022diagnosing}. Whenever we can find encoder and decoder functions $\psi,\phi$, respectively, such that $\phi(\psi(x))=x,\mu^A$ --- almost surely, and $\phi(\psi(x))=x,\mu^B$ --- almost surely, the \gls{KL} divergence can be computed in the \textit{latent} space obtained by the encoder $\psi$. Considering the pushforward measure ${\mu}^{A}\circ \psi^{-1}$, it is indeed possible to show (proof in \Cref{sec:proof_kl_deriv}) that $\KL{\mu^A}{\mu^B}=\KL{{\mu}^{A}\circ \psi^{-1}}{{\mu}^{B}\circ \psi^{-1}}$. This property is particularly useful as it allows using score based models trained in a latent space to compute the \gls{KL} divergences of interest, as we do in \Cref{sec:consistency}.

\subsection{\gls{KL} Estimators and Theoretical Guarantees}\label{sec:estimators}

Given the parametric approximations of the score networks through minimization of \Cref{eq:loss}, and the result in \Cref{kl_formula}, we are ready to discuss our proposed \textbf{estimator} of the \gls{KL} divergence. We focus on the first term on the r.h.s. of \Cref{kl_formula}, which has unknown value, and define its approximated version
\begin{equation}\label{eq:estimator}
    e(\mu^A,\mu^B)\defeq\E_{   {\mathbb{P}}^{\mu^A}}\left[\int\limits_0^T \frac{g^2_{t}}{2}\norm{\tilde{s}^{\mu^A}_t(X_t)-\tilde{s}^{\mu^B}_t(X_t) }^2\dd t\right]=\int\limits_0^T\frac{g^2_{t}}{2} \E_{\nu^{\mu^A}_t}\left[\norm{\tilde{s}^{\mu^A}_t(X_t)-\tilde{s}^{\mu^B}_t(X_t) }^2\right]\dd t,
\end{equation}
where parametric scores, instead of true score functions, are used. By defining the score error as $\epsilon^{\mu^A}_t(x)\defeq \tilde{s}^{\mu^A}_t(x)-s^{\mu^A}_t(x)$, it is possible to show (see \Cref{sec:proof_kl_deriv}) that $e(\mu^A,\mu^B)-\E_{   {\mathbb{P}}^{\mu^A}}\left[\int\limits_0^T\frac{g^2_{t}}{2}\norm{s^{\mu^A}_t(X_t)-s^{\mu^B}_t(X_t) }^2\dd t\right]$ has expression
\begin{flalign}\label{error_score}
  &d=\E_{   {\mathbb{P}}^{\mu^A}}\left[\int\limits_0^T\frac{g^2_{t}}{2}\norm{\epsilon^{\mu^A}_t(X_t)-\epsilon^{\mu^B}_t(X_t)}^2+2\langle {s}^{\mu^A}_t(X_t)-{s}^{\mu^B}_t(X_t),\epsilon^{\mu^A}_t(X_t)-\epsilon^{\mu^B}_t(X_t)\rangle\dd t\right].
\end{flalign}
As for the second term on the r.h.s. of \Cref{kl_formula}, $\KL{\nu^{\mu^A}_T}{\nu^{\mu^B}_T}$, we recall that it is a quantity that vanishes with large $T$. Consequently, given a sufficiently large diffusion time $T$ the function $e$ serves as an accurate estimator of the true \gls{KL}: 
\begin{equation}\label{estim_score}
    e(\mu^A,\mu^B)= \KL{\mu^A}{\mu^B}+d-\KL{\nu^{\mu^A}_T}{\nu^{\mu^B}_T}\simeq \KL{\mu^A}{\mu^B}.
\end{equation}
An important property of our estimator is that it is \textit{neither} an upper nor a lower bound of the true KL divergence: indeed the $d$ term of \Cref{estim_score} can be either positive or negative. This property, frees our estimation guarantees from the pessimistic results of \cite{mcallester2020formal}. Note also that, counter-intuitively, larger errors norms $\norm{\epsilon^{\mu^A}_t(x)}$ not necessarily imply larger estimation error of the \gls{KL} divergence. Indeed, common mode errors (reminiscent of paired statistical tests) cancel out. In the special case where $\epsilon^{\mu^A}_t(x)=\epsilon^{\mu^B}_t(x)$, the estimation error due to the approximate nature of the score functions is indeed zero.

Accurate quantification of the estimation error is, in general, a challenging task. Indeed, techniques akin to the works \citep{de2022convergence,lee2022convergence,chen2022sampling}, where guarantees are provided w.r.t. to the distance between the real backward dynamics and the measures induced by the \textit{simulated} backward dynamics, $\KL{\mu^{A}}{\tilde{\mu}^{A}}$, are not readily available in our context. Qualitatively, we observe that our estimator is affected by two sources of error: score networks that only approximate the true score function and finiteness of $T$.
The $d$ term in \Cref{estim_score}, which is related to the score discrepancy, suggests selection of a small time $T$ (indeed we can expect such mismatch to behave as a quantity that increases with $T$ \citep{franzese2022}). It is important however to adopt a sufficiently large diffusion time $T$ such that $\KL{\nu^{\mu^A}_T}{\nu^{\mu^B}_T}$ is negligible. Typical diffusion schedules satisfy these requirements.
Note that, if the \gls{KL} term is known (or approximately known), it can be included in the definition of the estimator function, reducing the estimation error (see also discussion in \Cref{sec:entropyest}).

\paragraph{Montecarlo Integration}
The analytical computation of \Cref{eq:estimator} is, in general, out of reach. However, Montecarlo integration is possible, by recognizing that samples from $\nu^{\mu^A}_t$ can be obtained through the sampling scheme ${   X_0\sim\mu^A,X_t\sim\nu_t^{\delta_{X_0}}}$. The outer integration w.r.t. to the time instant is similarly possible by sampling $t\sim\mathcal{U}(0,T)$, and multiplying the result of the estimation by $T$ (since $\int_0^T(\cdot) \dd t=T\E_{t\sim\mathcal{U}(0,T)}[(\cdot)]$). Alternatively, it is possible to implement importance sampling schemes to reduce the variance, along the lines of what described by \citet{huang2021variational}, by sampling the time instant non-uniformly and modifying accordingly the time-varying constants in \Cref{eq:estimator}. 
In both cases, the Montecarlo estimation error can be reduced to arbitrary small values by collecting enough samples, with guarantees described in \citep{rainforth2018nesting}.

\subsection{Entropy estimation}\label{sec:entropyest}

We now describe how to compute the entropy associated to a given density $\mu^A$, $\textsc{H}(\mu^A)\defeq\int \dd\mu^A(x)\log \bar{\mu}^A(x)$. Using the ideas for estimating the \gls{KL} divergence, we notice that we can compute $\KL{\mu^A}{\GaussM{\sigma}}$, where $\Gauss{\sigma}(x)$ stands for the standard Gaussian distribution with mean $0$ and covariance $\sigma^2 I$.
Then, we can relate the entropy to such divergence through the following equality:
\begin{flalign}\label{eq:kl_h}    
    \textsc{H}(\mu^A)+\KL{\mu^A}{\GaussM{\sigma}}=-\int \dd \mu^A(x)\log \Gauss{\sigma}(x)=\frac{N}{2}\log(2\pi \sigma^2)+\frac{\E_{\mu^A}\left[X_0^2\right]}{2\sigma^2}.
\end{flalign}

A simple manipulation of \Cref{eq:kl_h}, using the results from \Cref{sec:estimators}, implies that the estimation of the entropy $\textsc{H}(\mu^A)$ involves three unknown terms: $e(\mu^A,\GaussM{\sigma }),\KL{\nu^{\mu^A}_T}{\nu^{\GaussM{\sigma}}_T}$ and $\frac{\E_{\mu^A}\left[X_0^2\right]}{2\sigma^2}$. Now, the score function associated to the forward process starting from $\GaussM{\sigma}$ is analytically known and has value $s^{\GaussM{\sigma}}_t(x)= -\chi^{-1}_tx$, where
$\chi_t=\left(k^2_t\sigma^2+k^2_t\int_0^t k^{-2}_sg^2_{s}\dd s\right)I
$, with $k_t=\exp{\left(\int_0^tf_{s}\dd s\right)}$.
Moreover, whenever $T$ is large enough $\nu^{\mu^A}_T\simeq \GaussM{1}$, independently on the chosen value of $\sigma$. Consequently $\KL{\nu^{\mu^A}_T}{\nu^{\GaussM{\sigma}}_T}\simeq \KL{\GaussM{1}}{\GaussM{\sqrt{\chi_T}}}$, which is analytically available as $\nicefrac{N}{2}\left(\log{(\chi_{T}})-1+\nicefrac{1}{\chi_{T}}\right)$. Quantification of such approximation is possible following the same lines defined by \citet{collet2008logarithmic}. In summary, we consider the following estimator for the entropy:
\begin{equation}\label{entropy_est}
    {\textsc{H}}(\mu^A;\sigma)\simeq\frac{N}{2}\log(2\pi \sigma^2)+\frac{\E_{\mu^A}\left[X_0^2\right]}{2\sigma^2}-e(\mu^A,\GaussM{\sigma})-\frac{N}{2}\left(\log{(\chi_{T}})-1+\frac{1}{\chi_{T}}\right)
\end{equation}
For completeness, we note that a related estimator has recently appeared in the literature \citep{kong2023information}, although the technical derivation and objectives are different than ours.

\section{Computation of Mutual Information}\label{sec:mutualinfo}
In this work, we are interested in estimating the \gls{MI} between two random variables $A,B$. Consequently, we need to define the joint, conditional and marginal measures. We consider the first random variable $A$ in $\mathbb{R}^N$ to have marginal measure $\mu^{A}$. Similarly, we indicate the marginal measure of the second random variable $B$ with $\mu^{B}$. The joint measure of the two random variables $C\defeq[ A,B]$, which is defined in $\mathbb{R}^{2N}$, is indicated with $\mu^{C}$. What remains to be specified are the conditional measures of the first variable given a particular value of the second $A\g B= y$, shortened with $A_y$, that we indicate with the measure $\mu^{A_y}$, and the conditional measure of the second given a particular value of the first, $B\g A= x$, shortened with $B_x$, and indicated with $\mu^{B_x}$. This choice of notation, along with Bayes theorem, implies the following set of equivalences: $\dd\mu^{C}(x,y)=\dd\mu^{A_y}(x)\dd\mu^{B}(y)=\dd\mu^{B_x}(y)\dd\mu^{A}(x)$ and $\mu^{A}=\int \mu^{A_y}\dd\mu^{B}(y),\mu^{B}=\int\mu^{B_x}\dd\mu^{A}(x)$.

The marginal measures $\mu^A,\mu^B$ are associated to diffusion of the form of \Cref{eq:diffsdes}. Similarly, the joint $\mu^{C}$ and conditional $\mu^{A_y}$ measures we introduced, are associated to forward diffusion processes:
\begin{equation}\label{eq:diffsdesjoint}
    \begin{cases}
    \dd \left[X_t,Y_t\right]^\top =f_t[X_t,Y_t]^\top \dd t+ g_t\left[\dd W_t,\dd W'_t\right]^\top\\
    [X_0,Y_0]^\top\sim \mu^{C}
    \end{cases},\quad
    \begin{cases}
        \dd X_t=f_tX_t\dd t+ g_t\dd W_t \\
        X_0\sim \mu^{A_y}
    \end{cases}
\end{equation}
respectively, where the \gls{SDE} on the l.h.s. is valid for the real space $\mathbb{R}^{2N}$, as defined in \Cref{sec:diffusion}.

In this work, we consider two classes of diffusion processes. In the first case, the diffusion model is asymmetric, and the random variable $B$ is only considered as a conditioning signal. As such, we learn the score associated to the random variable $A$, with a conditioning signal $B$, which is set to some predefined null value when considering the marginal measure. This well-known approach \citep{ho2021classifier} effectively models the marginal and conditional scores associated to $\mu^A$ and $\mu^{A_y}$ with a unique score network. 

Next, we define a new kind of diffusion model for the joint random variable $C$, which allows modelling the joint and the conditional measures. Inspired by recent trends in multi-modal generative modeling \citep{bao2023transformer, bounoua2023multimodal}, we define a joint diffusion process that allows amortized training of a single score network, instead of considering separate diffusion processes and their respective score networks, for each random variable.
To do so, we define the following \gls{SDE}: 
\begin{equation}\label{eq:maskeddiffusion}
    \begin{cases}
    \dd \left[X_t,Y_t\right]^\top =f_t[\alpha X_t,\beta Y_t]^\top \dd t+ g_t\left[\alpha \dd W_t,\beta \dd W'_t\right]^\top,\\
    [X_0,Y_0]^\top\sim \mu^{C},
    \end{cases}
\end{equation}

with extra parameters $\alpha,\beta\in \{0,1\}$.
This \gls{SDE} extends the l.h.s. of \Cref{eq:diffsdesjoint}, and describes the joint evolution of the variables $X_t,Y_t$, starting from the joint measure $\mu^{C}$, with overall path measure $\mathbb{P}^{\mu^{C}}$. The two extra coefficients $\alpha,\beta$ are used to modulate the \textit{speed} at which the two portions $X_t,Y_t$ of the process diffuse towards their steady state. More precisely, $\alpha=\beta=1$ corresponds to a \textit{classical} simultaneous diffusion (l.h.s. of \Cref{eq:diffsdesjoint}).
On the other hand, the configuration $\alpha=1,\beta=0$ corresponds to the case in which the variable $Y_t$ remains constant throughout all the diffusion (which is used for conditional measures, r.h.s. of \Cref{eq:diffsdesjoint}). The specular case, $\alpha=0,\beta=1$, similarly allows to study the evolution of $Y_t$ conditioned on a constant value of $X_0$.
Then, instead of learning three separate score networks (for $\mu^C,\mu^{A_y}$ and $\mu^{B_x}$), associated to standard diffusion processes, the key idea is to consider a \textit{unique} parametric score, leveraging the unified formulation \Cref{eq:maskeddiffusion}, which accepts as inputs two vectors in $\mathbb{R}^{N}$, the diffusion time $t$, and the two coefficients $\alpha,\beta$. This allows to conflate in a single architecture: i) the score $s^{\mu^{C}}_{t}(x,y)$ associated to the joint diffusion of the variables $A,B$ (corresponding to $\alpha=\beta=1$) and ii) the conditional score $s^{\mu^{A_y}}_{t}(x)$ (corresponding to $\alpha=1,\beta=0$). 
Additional details are presented in \cref{apdx:imp_detail}.

\subsection{\gls{MINDE}: a Family of \gls{MI} estimators}

We are now ready to describe our new \gls{MI} estimator, which we call \gls{MINDE}. As a starting point, we recognize that the \gls{MI} between two random variables $A,B$ has several equivalent expressions, among which \Cref{eq:mi_1,eq:mi_2,eq:mi_3}. On the left hand side of these expressions we report well-known formulations for the \gls{MI}, $\textsc{I}(A,B)$, while on the right hand side we express them using the estimators we introduce in this work, where equality is assumed to be valid up to the errors described in \Cref{sec:kl_deriv}.  

\begin{footnotesize}
\begin{align}
    \textsc{H}(A)-\textsc{H}(A\g B)\simeq & -e(\mu^{A},\GaussM{\sigma})+\int e(\mu^{A_y},\GaussM{\sigma})\dd\mu^{B}(y)\label{eq:mi_2},\\
    \int \KL{\mu^{A_y}}{\mu^{A}} \dd\mu^{B}(y)\simeq & \int e(\mu^{A_y},\mu^{A})\dd\mu^{B}(y) \label{eq:mi_1},\\
    \textsc{H}(C)-\textsc{H}(A\g B)-\textsc{H}(B\g A)\simeq & -e(\mu^{C},\GaussM{\sigma})+\int e(\mu^{A_y},\GaussM{\sigma})\dd\mu^{B}(y)+\int e(\mu^{B_x},\GaussM{\sigma})\dd\mu^{A}(x).\label{eq:mi_3}
\end{align}
\end{footnotesize}
Note that it is possible to derive (details in \Cref{proof_eq:mi_3inner}) another equality for the \gls{MI}: 
\begin{equation}\label{eq:mi_3inner}
    \textsc{I}(A,B)\simeq\E_{\mathbb{P}^{\mu^{C}}}\left[\int\limits_0^T\frac{g^2_{t}}{2} \left[\norm{{\tilde s}^{\mu^{C}}_{t}([X_t,Y_t])-[\tilde s^{\mu^{A_{Y_0}}}_{t}(X_t),\tilde s^{\mu^{B_{X_0}}}_{t}(Y_t)] }^2\right]\dd t\right]. 
\end{equation}

Next, we describe how the conditional and joint modeling approaches can be leveraged to compute a \textit{family} of techniques to estimate \gls{MI}. We evaluate all the variants in \Cref{sec:experiments}.

\paragraph{Conditional Diffusion Models.} We start by considering conditional models. 
A simple \gls{MI} estimator can be obtained considering \Cref{eq:mi_2}.
The entropy of $A$ can be estimated using \Cref{entropy_est}. Similarly, we can estimate the conditional entropy $\textsc{H}(A\g B)$ using the equality $\textsc{H}(A\g B)=\int \textsc{H}(A_y)\dd\mu^{B}(y)$, where the argument of the integral, $\textsc{H}(A_y)$, can be again obtained using \Cref{entropy_est}. Notice, that since $\E_{\mu^B(y)}\E_{\mu^{A_y}}\left[X_0^2\right]=\E_{\mu^A}\left[X_0^2\right]$, when substracting the estimators of $\textsc{H}(A)$ and $\textsc{H}(A\g B)$, all the terms but the estimator functions $e(\cdot)$ cancels out, leading to the equality in \Cref{eq:mi_2}. A second option is to simply use \Cref{eq:mi_1} and leverage \Cref{eq:estimator}.

\paragraph{Joint diffusion models.} 
Armed with the definition of a joint diffusion processes, and the corresponding score function, we now describe the basic ingredients that allow estimation of the \gls{MI}, according to various formulations. Using the joint score function $s^{\mu^{C}}_{t}([x,y])$, the estimation of the joint entropy $\textsc{H}(A, B)$ can be obtained with a straightforward application of \Cref{entropy_est}. Similarly, the conditional entropy $\textsc{H}(A\g B)=\int \textsc{H}(A_y)\dd\mu^{B}(y)$ can be computed using $s^{\mu^{A_y}}_{t}(x)$ to obtain the conditional score. Notice that $\textsc{H}(B\g A)$ is similarly obtained. Given the above formulations of the joint and conditional entropy, it is now easy to compute the \gls{MI} according to \Cref{eq:mi_3}, where we notice that, similarly to what discussed for conditional models, many of the terms in the different entropy estimations cancel out. Finally, it is possible to compute the \gls{MI} according to \Cref{eq:mi_3inner}. Interestingly, this formulation allows to eliminate the need for the parameter $\sigma$ of the entropy estimators, similarly to the \gls{MINDE} conditional variant, which shares this property as well (\Cref{eq:mi_2}).

\section{Experimental Validation}\label{sec:experiments}
We now evaluate the different estimators proposed in \Cref{sec:mutualinfo}. In particular, we study conditional and joint models (\gls{MINDE}\--\textsc{c} and \gls{MINDE}\--\textsc{j} respectively), and variants that exploit the difference between the parametric scores \textit{inside} the same norm ( \Cref{eq:mi_1,eq:mi_3inner}) or \textit{outside} it, adopting the difference of entropies representation along with Gaussian reference scores $s^{\GaussM{c}}$ (\Cref{eq:mi_2,eq:mi_3}). Summarizing, we refer to the different variants as \gls{MINDE}\--\textsc{c}($\sigma$), \gls{MINDE}\--\textsc{c}, and \gls{MINDE}\--\textsc{j}($\sigma$), \gls{MINDE}\--\textsc{j}, for \Cref{eq:mi_2,eq:mi_1,eq:mi_3,eq:mi_3inner} respectively. Our empirical validation involves a large range of synthetic distributions, which we present in \Cref{sec:benchmarks}. We also analyze the behavior of all \gls{MINDE} variants according to \textit{self-consistency} tests, as discussed in \Cref{sec:consistency}.

For all the settings, we use a simple, stacked multi-layer perception (MLP) with skip connections adapted to the input dimensions, and adopt  \textsc{vp}-\gls{SDE} diffusion \cite{song2021a}. We apply importance sampling \citep{huang2021variational,song2021a} at both training and inference time. More details about the implementation are included in \Cref{apdx:imp_detail}.

\subsection{\gls{MI} Estimation Benchmark}\label{sec:benchmarks}

We use the evaluation strategy proposed by \citet{czyz2023beyond}, which covers a range of distributions going beyond what is typically used to benchmark \gls{MI} estimators, e.g., multivariate normal distributions. In summary, we consider high-dimensional cases with (possibly) long-tailed distributions and/or sparse interactions, in the presence of several non trivial non-linear transformation. Benchmarks are constructed using samples from several base distributions, including Uniform, Normal with either dense or sparse correlation structure, and long-tailed Student distributions. Such samples are further modified by deterministic transformations, including the Half-Cube homeomorphism, which extends the distribution tails, and the Asinh Mapping, which instead shortens them, the Swiss Roll Embedding and Spiral diffeomorphis, which alter the simple linear structure of the base distributions. 

We compare \gls{MINDE} against neural estimators, such as \textsc{mine}~\citep{belghazi2018mine}, 
\textsc{Infonce}~\citep{oord2019representation},\textsc{nwj}~\citep{nguyen2007neurips} and \textsc{doe}~\citep{mcallester2020formal}. 
To ensure a fair comparison between \gls{MINDE} and other neural competitors, we consider architectures with a comparable number of parameters. Note that the original benchmark in \citep{czyz2023beyond} uses 10k training samples, which are in many cases not sufficient to obtain stable estimates of the \gls{MI} for our competitors. Here, we use a larger training size (100k samples) to avoid confounding factors in our analysis. 
In all our experiments, we fix $ \sigma = 1.0 $ for the \gls{MINDE}\--\textsc{c}($\sigma$), \gls{MINDE}\--\textsc{j}($\sigma$) variants, which results in the best performance (an ablation study is included in \Cref{apdx:sigma_ablations}).

\paragraph{Results:}
The general benchmark consists of 40 tasks (10 unique tasks$\times$ 4 parametrizations) designed  by combining distributions and MI-invariant transformations discussed earlier. 
We average results over 10 seeds for \gls{MINDE} variants and competitors, following the same protocol as in \cite{czyz2023beyond}. 
We present the full set of \gls{MI} estimation tasks in \Cref{tab:benchmark}. As in the original \cite{czyz2023beyond}, estimates for the different methods are presented with a precision of 0.1 nats, to improve visualization. For low-dimensional distributions, benchmark results show that all methods are effective in accurate \gls{MI} estimation. Differences emerge for more challenging scenarios. Overall, all our \gls{MINDE} variants perform well. \gls{MINDE}--\textsc{c} stands out as the best estimator with 35/40 estimated tasks with an error within the 0.1 nats quantization range.
Moreover, \gls{MINDE} can accurately estimate the \gls{MI} for long tailed distributions (Student) and highly transformed distributions (Spiral, Normal CDF), which are instead problematic for most of the other methods. The \textsc{mine} estimator achieves the second best performance, with an \gls{MI} estimation within 0.1 nats from ground truth for 24/40 tasks. Similarly to the other neural estimator baselines, \textsc{mine} is limited when dealing with long tail distributions (Student), and significantly transformed distributions (Spiral).

 \paragraph{High MI benchmark:}
 
\begin{figure}[ht]
    \centering
    \begin{subfigure}{0.25\textwidth}
        \includegraphics[page=1,width=\linewidth]{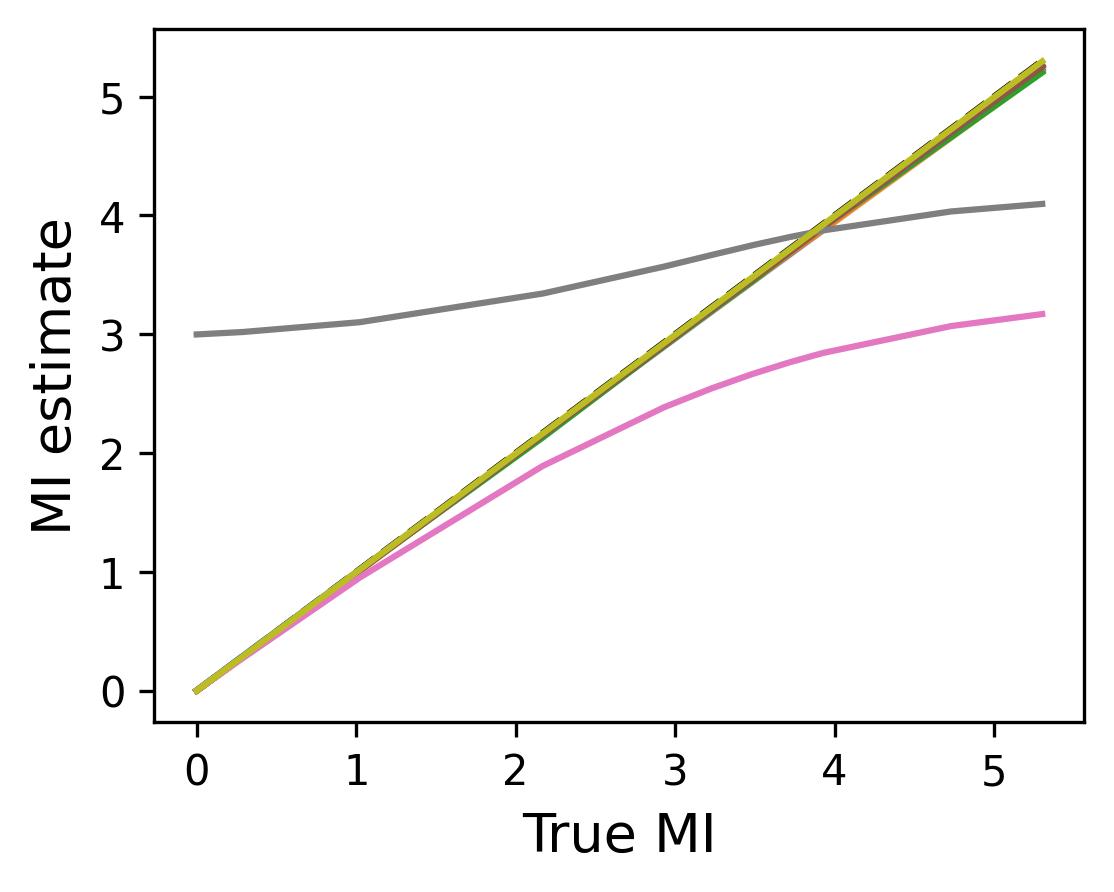}
        \caption{ Sparse Multinormal}
    \end{subfigure}
    \begin{subfigure}{0.25\textwidth}
        \includegraphics[page=1,width=\linewidth]{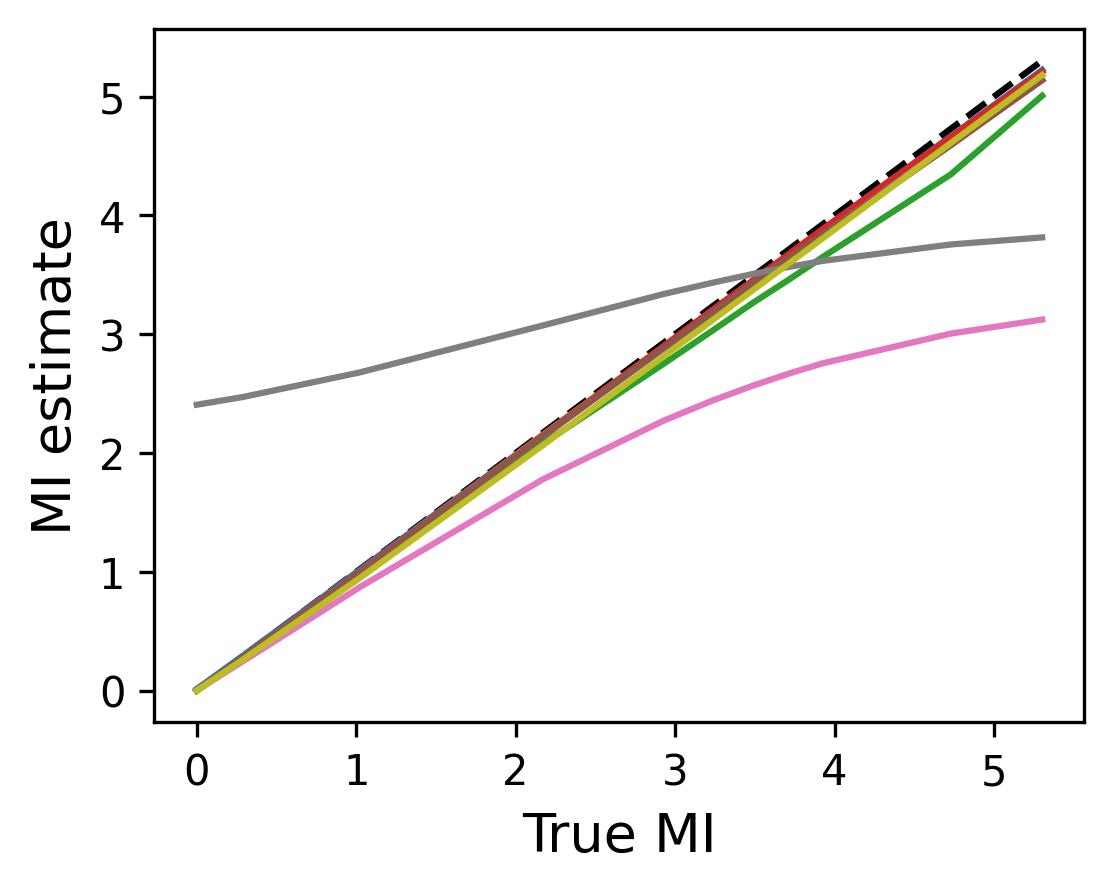}
        \caption{Half-cube}
    \end{subfigure}
      \begin{subfigure}{0.25\textwidth}
        \includegraphics[page=1,width=\linewidth]{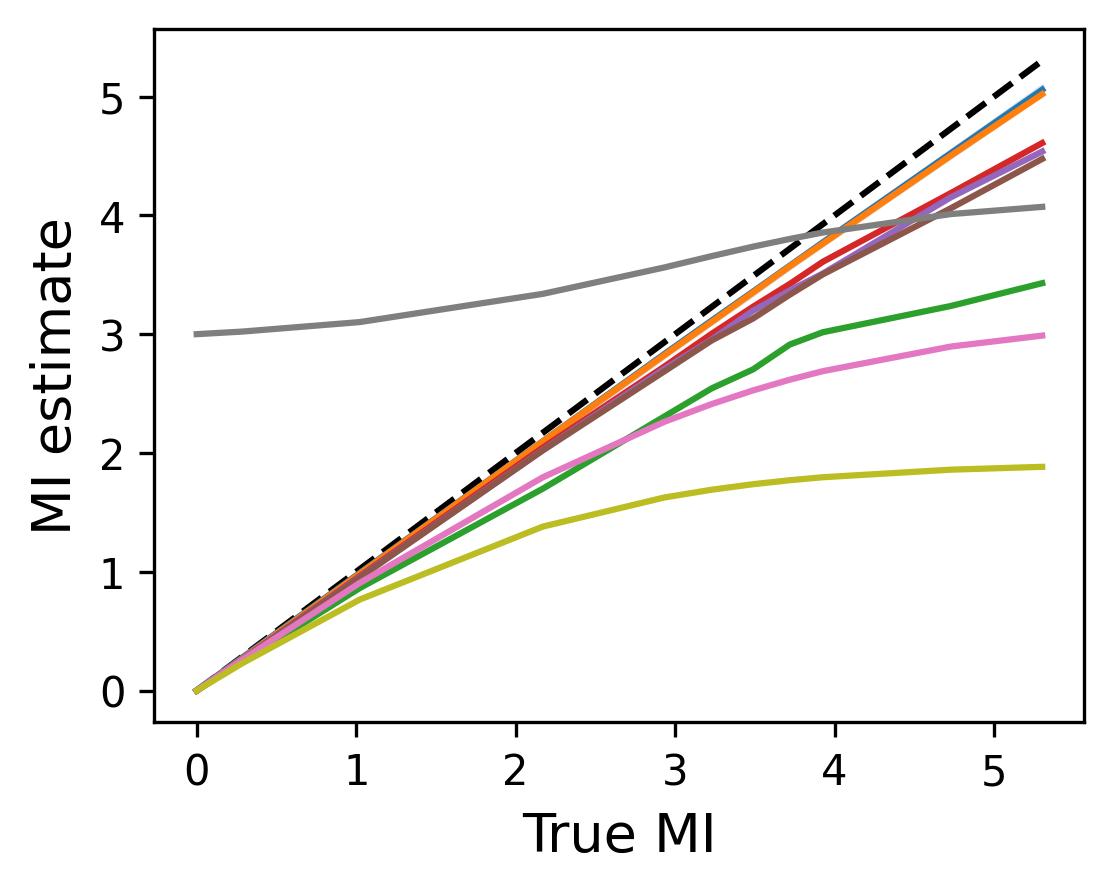}
        \caption{Spiral  }
    
    \end{subfigure}
    \begin{subfigure}{0.1\textwidth}
        \includegraphics[page=1,width=\linewidth]{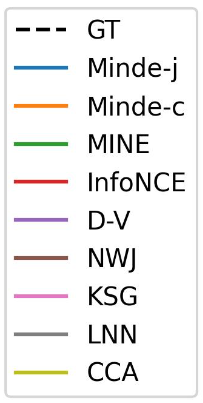}
      
    \end{subfigure}
    \caption{High \gls{MI} benchmark: original (column (a)) and transformed variants (columns (b) and (c)). 
     }
    \label{fig:high_mi}
\end{figure}

Through this second benchmark, we target high \gls{MI} distributions. We consider $3\cross3$  multivariate normal distribution with sparse interactions as done in \cite{czyz2023beyond}. We vary the correlation parameter to obtain the desired \gls{MI}, and test the estimators when applying Half-cube or Spiral transformations. Results in \Cref{fig:high_mi} show that while on the non transformed distribution (column (a)) all neural estimators nicely follow the ground truth, on the transformed versions (columns (b) and (c)), \gls{MINDE} outperforms competitors.
 
\vspace*{-2cm}
\subsection{Consistency tests}
\label{sec:consistency}

The second set of tests we perform are the self-consistency ones proposed in \cite{song2019understanding}, which aim at investigating properties of \gls{MI} estimators on real data. Considering as random variable $A$ a sample from the \textsc{mnist} (resolution $28\times28$) dataset, the first set of measurements performed is the estimation of $\textsc{I}(A,B_r)$, where $B_r$ is equal to $A$ for the first $r$ rows, and set to 0 afterwards. It is evident that $\textsc{I}(A,B_r)$ is a quantity that increases with $r$, where in particular $\textsc{I}(A,B_0)=0$. Testing whether this holds also for the estimated \gls{MI} is referred to as \textit{independency} test. The second test proposed in \cite{song2019understanding} is the \textit{data-processing} test, where given that $\textsc{I}(A;[B_{r+k},B_r])=\textsc{I}(A;B_{r+k}),k>0$, the task is to verify it through estimators for different values of $k$. Finally, the \textit{additivity} tests aim at assessing whether for two independent images $A^1,A^2$ extracted from the dataset, the property $\textsc{I}([A^1,A^2];[B^1_{r},B^2_r])=2{\textsc{I}}(A^1;B^1_r)$ is satisfied also by the numerical estimations.

\begin{figure}[ht]
    \centering
        
    \begin{subfigure}{0.25\textwidth}
        \includegraphics[page=1,width=\linewidth]{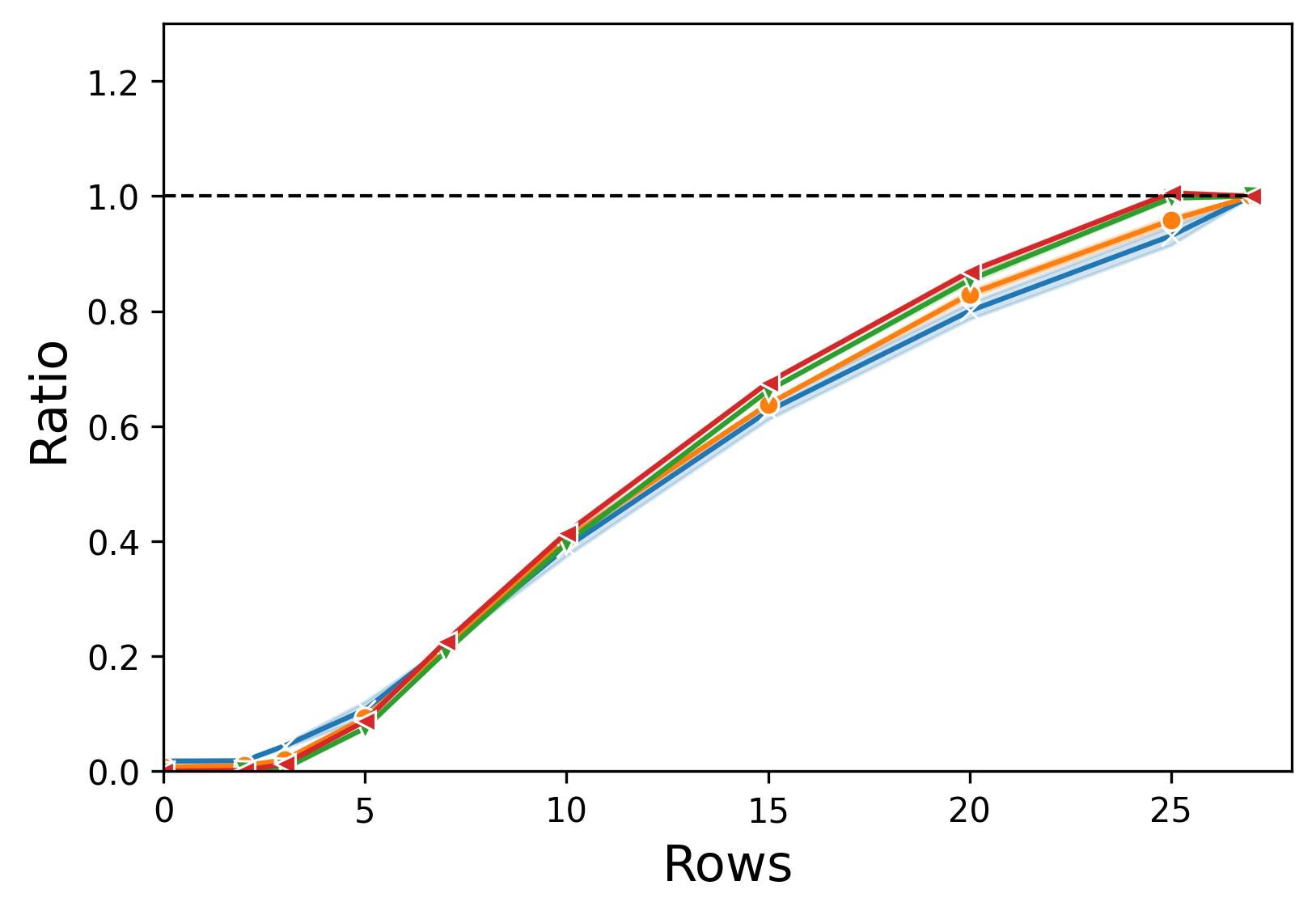}
        \caption{ Baseline test }
    \label{fig:set_1}\end{subfigure}
    \begin{subfigure}{0.25\textwidth}
        \includegraphics[page=1,width=\linewidth]{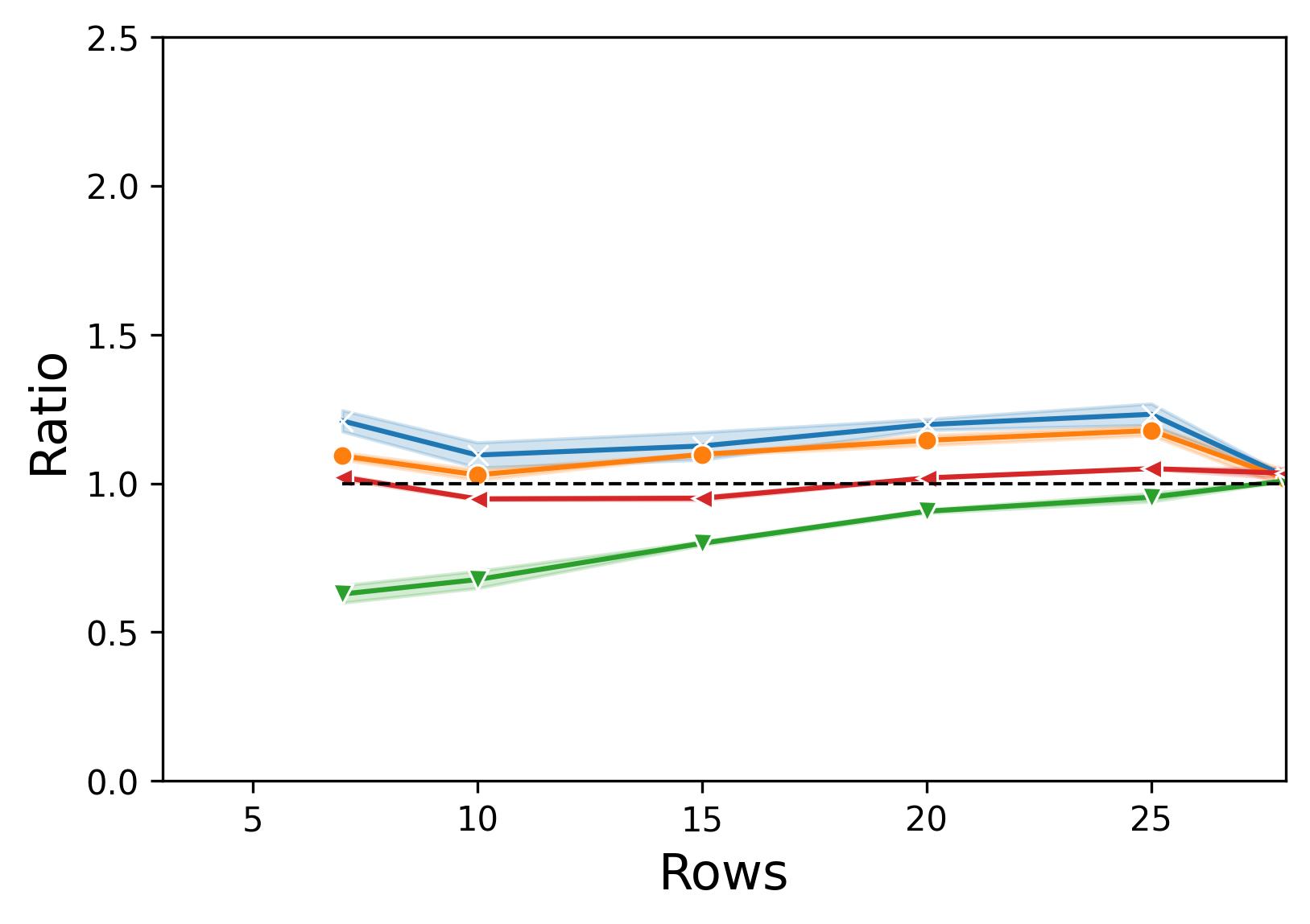}
        \caption{Data processing test }
    \label{fig:set_2}
    \end{subfigure}
      \begin{subfigure}{0.25\textwidth}
        \includegraphics[page=1,width=\linewidth]{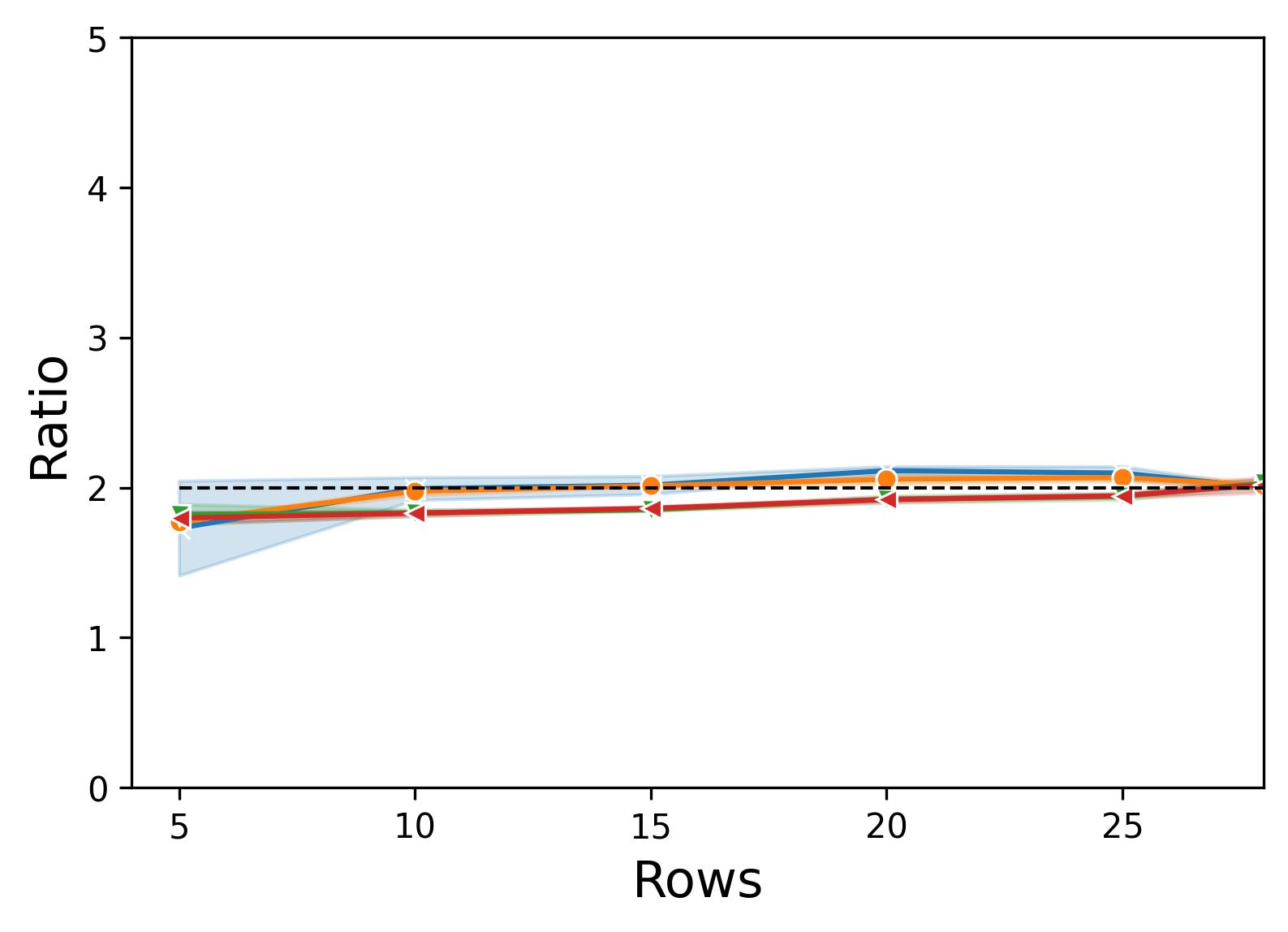}
        \caption{Additivity test  }
    \label{fig:set_3}
    \end{subfigure}
    \begin{subfigure}{0.15\textwidth}
\includegraphics[page=1,width=\linewidth]{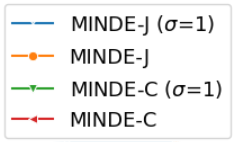}
          \caption*{  }
        \end{subfigure}
        
    \caption{Consistency tests results on the \textsc{mnist} dataset. \textit{ Baseline test \Cref{fig:set_1}:}  Evaluation of $ \frac{I(A,B_r)}{I(A,B_0)}$. $A$ is an image and $B_r$ is an image containing the top $t$ rows of $A$. 
    \textit{ Data processing test \Cref{fig:set_2}:} Evaluation of $ \frac{I(A,[B_{r+k},B_{r})] )}{I(A,B_{r+k})}$ (ideal value is 1). 
     \textit{ Additivity test \Cref{fig:set_3}:}  Evaluation of $ \frac{I( [A^1, A^2],[B^1_r, B^2_r] )}{I(A^1,B^1_r)}$ (ideal value is 2).    
     }
    \label{fig:consistency_rests}
\end{figure}

For these tests, we consider diffusion models in a latent space, exploiting the invariance of \gls{KL} divergences to perfect auto-encoding (see \Cref{sec:kl_deriv}). First, we train for all tests deterministic auto-encoders for the considered images. Then, through concatenation of the latent variables, as done in \citep{bao2023transformer, bounoua2023multimodal}, we compute the \gls{MI} with the different schemes proposed in \Cref{sec:mutualinfo}. Results of the three tests (averaged over 5 seeds) are reported in
\Cref{fig:consistency_rests}. In general, all \gls{MINDE} variants show excellent performance, whereas none of the other neural \gls{MI} estimators succeed at passing simultaneously all tests, as can be observed from Figures 4,5,6 in the original \cite{song2019understanding}).

\renewcommand{\tabcolsep}{0.0pt}

\begin{table}[H]
\vspace*{-0cm}
\tiny

\resizebox{\textwidth}{!}{
\begin{tabular}{lrrrrrrrrrrrrrrrrrrrrrrrrrrrrrrrrrrrrrrrr}
\toprule

GT & 0.2 & 0.4 & 0.3 & 0.4 & 0.4 & 0.4 & 0.4 & 1.0 & 1.0 & 1.0 & 1.0 & 0.3 & 1.0 & 1.3 & 1.0 & 0.4 & 1.0 & 0.6 & 1.6 & 0.4 & 1.0 & 1.0 & 1.0 & 1.0 & 1.0 & 1.0 & 1.0 & 1.0 & 1.0 & 0.2 & 0.4 & 0.2 & 0.3 & 0.2 & 0.4 & 0.3 & 0.4 & 1.7 & 0.3 & 0.4 \\
\midrule

\textbf{\gls{MINDE}--\textsc{j} ($\sigma=1$)} \ & {\cellcolor[HTML]{F2F2F2}} 0.2 & {\cellcolor[HTML]{F2F2F2}} 0.4 & {\cellcolor[HTML]{F2F2F2}} 0.3 & {\cellcolor[HTML]{F2F2F2}} 0.4 & {\cellcolor[HTML]{F2F2F2}} 0.4 & {\cellcolor[HTML]{F2F2F2}} 0.4 & {\cellcolor[HTML]{F2F2F2}} 0.4 & {\cellcolor[HTML]{D4DDE9}} 1.1 & {\cellcolor[HTML]{F2F2F2}} 1.0 & {\cellcolor[HTML]{F2F2F2}} 1.0 & {\cellcolor[HTML]{F2F2F2}} 1.0 & {\cellcolor[HTML]{F2F2F2}} 0.3 & {\cellcolor[HTML]{EED3DB}} 0.9 & {\cellcolor[HTML]{EED3DB}} 1.2 & {\cellcolor[HTML]{F2F2F2}} 1.0 & {\cellcolor[HTML]{F2F2F2}} 0.4 & {\cellcolor[HTML]{F2F2F2}} 1.0 & {\cellcolor[HTML]{F2F2F2}} 0.6 & {\cellcolor[HTML]{D4DDE9}} 1.7 & {\cellcolor[HTML]{F2F2F2}} 0.4 & {\cellcolor[HTML]{F2F2F2}} 1.0 & {\cellcolor[HTML]{F2F2F2}} 1.0 & {\cellcolor[HTML]{F2F2F2}} 1.0 & {\cellcolor[HTML]{EED3DB}} 0.9 & {\cellcolor[HTML]{EED3DB}} 0.9 & {\cellcolor[HTML]{EED3DB}} 0.9 & {\cellcolor[HTML]{F2F2F2}} 1.0 & {\cellcolor[HTML]{EED3DB}} 0.9 & {\cellcolor[HTML]{F2F2F2}} 1.0 & {\cellcolor[HTML]{F2F2F2}} 0.2 & {\cellcolor[HTML]{F2F2F2}} 0.4 & {\cellcolor[HTML]{F2F2F2}} 0.2 & {\cellcolor[HTML]{F2F2F2}} 0.3 & {\cellcolor[HTML]{F2F2F2}} 0.2 & {\cellcolor[HTML]{D4DDE9}} 0.5 & {\cellcolor[HTML]{F2F2F2}} 0.3 & {\cellcolor[HTML]{D4DDE9}} 0.5 & {\cellcolor[HTML]{EED3DB}} 1.6 & {\cellcolor[HTML]{F2F2F2}} 0.3 & {\cellcolor[HTML]{F2F2F2}} 0.4 \\

\textbf{\gls{MINDE}--\textsc{j}}  & {\cellcolor[HTML]{F2F2F2}} 0.2 & {\cellcolor[HTML]{F2F2F2}} 0.4 & {\cellcolor[HTML]{F2F2F2}} 0.3 & {\cellcolor[HTML]{F2F2F2}} 0.4 & {\cellcolor[HTML]{F2F2F2}} 0.4 & {\cellcolor[HTML]{F2F2F2}} 0.4 & {\cellcolor[HTML]{F2F2F2}} 0.4 & {\cellcolor[HTML]{AFC4DD}} 1.2 & {\cellcolor[HTML]{F2F2F2}} 1.0 & {\cellcolor[HTML]{F2F2F2}} 1.0 & {\cellcolor[HTML]{F2F2F2}} 1.0 & {\cellcolor[HTML]{F2F2F2}} 0.3 & {\cellcolor[HTML]{F2F2F2}} 1.0 & {\cellcolor[HTML]{F2F2F2}} 1.3 & {\cellcolor[HTML]{F2F2F2}} 1.0 & {\cellcolor[HTML]{F2F2F2}} 0.4 & {\cellcolor[HTML]{F2F2F2}} 1.0 & {\cellcolor[HTML]{F2F2F2}} 0.6 & {\cellcolor[HTML]{D4DDE9}} 1.7 & {\cellcolor[HTML]{F2F2F2}} 0.4 & {\cellcolor[HTML]{D4DDE9}} 1.1 & {\cellcolor[HTML]{F2F2F2}} 1.0 & {\cellcolor[HTML]{F2F2F2}} 1.0 & {\cellcolor[HTML]{F2F2F2}} 1.0 & {\cellcolor[HTML]{EED3DB}} 0.9 & {\cellcolor[HTML]{EED3DB}} 0.9 & {\cellcolor[HTML]{D4DDE9}} 1.1 & {\cellcolor[HTML]{F2F2F2}} 1.0 & {\cellcolor[HTML]{F2F2F2}} 1.0 & {\cellcolor[HTML]{EED3DB}} 0.1 & {\cellcolor[HTML]{E7ADBE}} 0.2 & {\cellcolor[HTML]{F2F2F2}} 0.2 & {\cellcolor[HTML]{F2F2F2}} 0.3 & {\cellcolor[HTML]{F2F2F2}} 0.2 & {\cellcolor[HTML]{D4DDE9}} 0.5 & {\cellcolor[HTML]{F2F2F2}} 0.3 & {\cellcolor[HTML]{F2F2F2}} 0.4 & {\cellcolor[HTML]{F2F2F2}} 1.7 & {\cellcolor[HTML]{F2F2F2}} 0.3 & {\cellcolor[HTML]{F2F2F2}} 0.4 \\

\textbf{\gls{MINDE}--\textsc{c} ($\sigma=1$)} & {\cellcolor[HTML]{F2F2F2}} 0.2 & {\cellcolor[HTML]{F2F2F2}} 0.4 & {\cellcolor[HTML]{F2F2F2}} 0.3 & {\cellcolor[HTML]{F2F2F2}} 0.4 & {\cellcolor[HTML]{F2F2F2}} 0.4 & {\cellcolor[HTML]{F2F2F2}} 0.4 & {\cellcolor[HTML]{F2F2F2}} 0.4 & {\cellcolor[HTML]{F2F2F2}} 1.0 & {\cellcolor[HTML]{F2F2F2}} 1.0 & {\cellcolor[HTML]{F2F2F2}} 1.0 & {\cellcolor[HTML]{F2F2F2}} 1.0 & {\cellcolor[HTML]{F2F2F2}} 0.3 & {\cellcolor[HTML]{F2F2F2}} 1.0 & {\cellcolor[HTML]{F2F2F2}} 1.3 & {\cellcolor[HTML]{F2F2F2}} 1.0 & {\cellcolor[HTML]{F2F2F2}} 0.4 & {\cellcolor[HTML]{F2F2F2}} 1.0 & {\cellcolor[HTML]{F2F2F2}} 0.6 & {\cellcolor[HTML]{F2F2F2}} 1.6 & {\cellcolor[HTML]{F2F2F2}} 0.4 & {\cellcolor[HTML]{EED9DF}} 0.9 & {\cellcolor[HTML]{F2F2F2}} 1.0 & {\cellcolor[HTML]{F2F2F2}} 1.0 & {\cellcolor[HTML]{EED9DF}} 0.9 & {\cellcolor[HTML]{EED9DF}} 0.9 & {\cellcolor[HTML]{EED9DF}} 0.9 & {\cellcolor[HTML]{EED9DF}} 0.9 & {\cellcolor[HTML]{F2F2F2}} 1.0 & {\cellcolor[HTML]{EED9DF}} 0.9 & {\cellcolor[HTML]{EED9DF}} 0.1 & {\cellcolor[HTML]{EED9DF}} 0.3 & {\cellcolor[HTML]{F2F2F2}} 0.2 & {\cellcolor[HTML]{F2F2F2}} 0.3 & {\cellcolor[HTML]{F2F2F2}} 0.2 & {\cellcolor[HTML]{F2F2F2}} 0.4 & {\cellcolor[HTML]{F2F2F2}} 0.3 & {\cellcolor[HTML]{EED9DF}} 0.3 & {\cellcolor[HTML]{F2F2F2}} 1.7 & {\cellcolor[HTML]{F2F2F2}} 0.3 & {\cellcolor[HTML]{F2F2F2}} 0.4 \\

\textbf{\gls{MINDE}--\textsc{c}} & {\cellcolor[HTML]{F2F2F2}} 0.2 & {\cellcolor[HTML]{F2F2F2}} 0.4 & {\cellcolor[HTML]{F2F2F2}} 0.3 & {\cellcolor[HTML]{F2F2F2}} 0.4 & {\cellcolor[HTML]{F2F2F2}} 0.4 & {\cellcolor[HTML]{F2F2F2}} 0.4 & {\cellcolor[HTML]{F2F2F2}} 0.4 & {\cellcolor[HTML]{F2F2F2}} 1.0 & {\cellcolor[HTML]{F2F2F2}} 1.0 & {\cellcolor[HTML]{F2F2F2}} 1.0 & {\cellcolor[HTML]{F2F2F2}} 1.0 & {\cellcolor[HTML]{F2F2F2}} 0.3 & {\cellcolor[HTML]{F2F2F2}} 1.0 & {\cellcolor[HTML]{F2F2F2}} 1.3 & {\cellcolor[HTML]{F2F2F2}} 1.0 & {\cellcolor[HTML]{F2F2F2}} 0.4 & {\cellcolor[HTML]{F2F2F2}} 1.0 & {\cellcolor[HTML]{F2F2F2}} 0.6 & {\cellcolor[HTML]{F2F2F2}} 1.6 & {\cellcolor[HTML]{F2F2F2}} 0.4 & {\cellcolor[HTML]{F2F2F2}} 1.0 & {\cellcolor[HTML]{F2F2F2}} 1.0 & {\cellcolor[HTML]{F2F2F2}} 1.0 & {\cellcolor[HTML]{EED9DF}} 0.9 & {\cellcolor[HTML]{EED9DF}} 0.9 & {\cellcolor[HTML]{EED9DF}} 0.9 & {\cellcolor[HTML]{F2F2F2}} 1.0 & {\cellcolor[HTML]{F2F2F2}} 1.0 & {\cellcolor[HTML]{F2F2F2}} 1.0 & {\cellcolor[HTML]{EED9DF}} 0.1 & {\cellcolor[HTML]{EED9DF}} 0.3 & {\cellcolor[HTML]{F2F2F2}} 0.2 & {\cellcolor[HTML]{F2F2F2}} 0.3 & {\cellcolor[HTML]{F2F2F2}} 0.2 & {\cellcolor[HTML]{F2F2F2}} 0.4 & {\cellcolor[HTML]{F2F2F2}} 0.3 & {\cellcolor[HTML]{F2F2F2}} 0.4 & {\cellcolor[HTML]{F2F2F2}} 1.7 & {\cellcolor[HTML]{F2F2F2}} 0.3 & {\cellcolor[HTML]{F2F2F2}} 0.4 \\

\midrule
MINE & {\cellcolor[HTML]{F2F2F2}} 0.2 & {\cellcolor[HTML]{F2F2F2}} 0.4 & {\cellcolor[HTML]{EED3DB}} 0.2 & {\cellcolor[HTML]{F2F2F2}} 0.4 & {\cellcolor[HTML]{F2F2F2}} 0.4 & {\cellcolor[HTML]{F2F2F2}} 0.4 & {\cellcolor[HTML]{F2F2F2}} 0.4 & {\cellcolor[HTML]{F2F2F2}} 1.0 & {\cellcolor[HTML]{F2F2F2}} 1.0 & {\cellcolor[HTML]{F2F2F2}} 1.0 & {\cellcolor[HTML]{F2F2F2}} 1.0 & {\cellcolor[HTML]{F2F2F2}} 0.3 & {\cellcolor[HTML]{F2F2F2}} 1.0 & {\cellcolor[HTML]{F2F2F2}} 1.3 & {\cellcolor[HTML]{F2F2F2}} 1.0 & {\cellcolor[HTML]{F2F2F2}} 0.4 & {\cellcolor[HTML]{F2F2F2}} 1.0 & {\cellcolor[HTML]{F2F2F2}} 0.6 & {\cellcolor[HTML]{F2F2F2}} 1.6 & {\cellcolor[HTML]{F2F2F2}} 0.4 & {\cellcolor[HTML]{EED3DB}} 0.9 & {\cellcolor[HTML]{EED3DB}} 0.9 & {\cellcolor[HTML]{EED3DB}} 0.9 & {\cellcolor[HTML]{E7ADBE}} 0.8 & {\cellcolor[HTML]{E188A2}} 0.7 & {\cellcolor[HTML]{DB6185}} 0.6 & {\cellcolor[HTML]{EED3DB}} 0.9 & {\cellcolor[HTML]{EED3DB}} 0.9 & {\cellcolor[HTML]{EED3DB}} 0.9 & {\cellcolor[HTML]{E7ADBE}} 0.0 & {\cellcolor[HTML]{DB6185}} 0.0 & {\cellcolor[HTML]{EED3DB}} 0.1 & {\cellcolor[HTML]{E7ADBE}} 0.1 & {\cellcolor[HTML]{EED3DB}} 0.1 & {\cellcolor[HTML]{E7ADBE}} 0.2 & {\cellcolor[HTML]{EED3DB}} 0.2 & {\cellcolor[HTML]{F2F2F2}} 0.4 & {\cellcolor[HTML]{F2F2F2}} 1.7 & {\cellcolor[HTML]{F2F2F2}} 0.3 & {\cellcolor[HTML]{F2F2F2}} 0.4 \\
InfoNCE & {\cellcolor[HTML]{F2F2F2}} 0.2 & {\cellcolor[HTML]{F2F2F2}} 0.4 & {\cellcolor[HTML]{F2F2F2}} 0.3 & {\cellcolor[HTML]{F2F2F2}} 0.4 & {\cellcolor[HTML]{F2F2F2}} 0.4 & {\cellcolor[HTML]{F2F2F2}} 0.4 & {\cellcolor[HTML]{F2F2F2}} 0.4 & {\cellcolor[HTML]{F2F2F2}} 1.0 & {\cellcolor[HTML]{F2F2F2}} 1.0 & {\cellcolor[HTML]{F2F2F2}} 1.0 & {\cellcolor[HTML]{F2F2F2}} 1.0 & {\cellcolor[HTML]{F2F2F2}} 0.3 & {\cellcolor[HTML]{F2F2F2}} 1.0 & {\cellcolor[HTML]{F2F2F2}} 1.3 & {\cellcolor[HTML]{F2F2F2}} 1.0 & {\cellcolor[HTML]{F2F2F2}} 0.4 & {\cellcolor[HTML]{F2F2F2}} 1.0 & {\cellcolor[HTML]{F2F2F2}} 0.6 & {\cellcolor[HTML]{F2F2F2}} 1.6 & {\cellcolor[HTML]{F2F2F2}} 0.4 & {\cellcolor[HTML]{EED3DB}} 0.9 & {\cellcolor[HTML]{F2F2F2}} 1.0 & {\cellcolor[HTML]{F2F2F2}} 1.0 & {\cellcolor[HTML]{E7ADBE}} 0.8 & {\cellcolor[HTML]{E7ADBE}} 0.8 & {\cellcolor[HTML]{E7ADBE}} 0.8 & {\cellcolor[HTML]{EED3DB}} 0.9 & {\cellcolor[HTML]{F2F2F2}} 1.0 & {\cellcolor[HTML]{F2F2F2}} 1.0 & {\cellcolor[HTML]{F2F2F2}} 0.2 & {\cellcolor[HTML]{EED3DB}} 0.3 & {\cellcolor[HTML]{F2F2F2}} 0.2 & {\cellcolor[HTML]{F2F2F2}} 0.3 & {\cellcolor[HTML]{F2F2F2}} 0.2 & {\cellcolor[HTML]{F2F2F2}} 0.4 & {\cellcolor[HTML]{F2F2F2}} 0.3 & {\cellcolor[HTML]{F2F2F2}} 0.4 & {\cellcolor[HTML]{F2F2F2}} 1.7 & {\cellcolor[HTML]{F2F2F2}} 0.3 & {\cellcolor[HTML]{F2F2F2}} 0.4 \\
D-V & {\cellcolor[HTML]{F2F2F2}} 0.2 & {\cellcolor[HTML]{F2F2F2}} 0.4 & {\cellcolor[HTML]{F2F2F2}} 0.3 & {\cellcolor[HTML]{F2F2F2}} 0.4 & {\cellcolor[HTML]{F2F2F2}} 0.4 & {\cellcolor[HTML]{F2F2F2}} 0.4 & {\cellcolor[HTML]{F2F2F2}} 0.4 & {\cellcolor[HTML]{F2F2F2}} 1.0 & {\cellcolor[HTML]{F2F2F2}} 1.0 & {\cellcolor[HTML]{F2F2F2}} 1.0 & {\cellcolor[HTML]{F2F2F2}} 1.0 & {\cellcolor[HTML]{F2F2F2}} 0.3 & {\cellcolor[HTML]{F2F2F2}} 1.0 & {\cellcolor[HTML]{F2F2F2}} 1.3 & {\cellcolor[HTML]{F2F2F2}} 1.0 & {\cellcolor[HTML]{F2F2F2}} 0.4 & {\cellcolor[HTML]{F2F2F2}} 1.0 & {\cellcolor[HTML]{F2F2F2}} 0.6 & {\cellcolor[HTML]{F2F2F2}} 1.6 & {\cellcolor[HTML]{F2F2F2}} 0.4 & {\cellcolor[HTML]{EED3DB}} 0.9 & {\cellcolor[HTML]{F2F2F2}} 1.0 & {\cellcolor[HTML]{F2F2F2}} 1.0 & {\cellcolor[HTML]{E7ADBE}} 0.8 & {\cellcolor[HTML]{E7ADBE}} 0.8 & {\cellcolor[HTML]{E7ADBE}} 0.8 & {\cellcolor[HTML]{EED3DB}} 0.9 & {\cellcolor[HTML]{F2F2F2}} 1.0 & {\cellcolor[HTML]{F2F2F2}} 1.0 & {\cellcolor[HTML]{E7ADBE}} 0.0 & {\cellcolor[HTML]{DB6185}} 0.0 & {\cellcolor[HTML]{EED3DB}} 0.1 & {\cellcolor[HTML]{E7ADBE}} 0.1 & {\cellcolor[HTML]{F2F2F2}} 0.2 & {\cellcolor[HTML]{E7ADBE}} 0.2 & {\cellcolor[HTML]{EED3DB}} 0.2 & {\cellcolor[HTML]{F2F2F2}} 0.4 & {\cellcolor[HTML]{F2F2F2}} 1.7 & {\cellcolor[HTML]{F2F2F2}} 0.3 & {\cellcolor[HTML]{F2F2F2}} 0.4 \\
NWJ & {\cellcolor[HTML]{F2F2F2}} 0.2 & {\cellcolor[HTML]{F2F2F2}} 0.4 & {\cellcolor[HTML]{F2F2F2}} 0.3 & {\cellcolor[HTML]{F2F2F2}} 0.4 & {\cellcolor[HTML]{F2F2F2}} 0.4 & {\cellcolor[HTML]{F2F2F2}} 0.4 & {\cellcolor[HTML]{F2F2F2}} 0.4 & {\cellcolor[HTML]{F2F2F2}} 1.0 & {\cellcolor[HTML]{F2F2F2}} 1.0 & {\cellcolor[HTML]{F2F2F2}} 1.0 & {\cellcolor[HTML]{F2F2F2}} 1.0 & {\cellcolor[HTML]{F2F2F2}} 0.3 & {\cellcolor[HTML]{F2F2F2}} 1.0 & {\cellcolor[HTML]{F2F2F2}} 1.3 & {\cellcolor[HTML]{F2F2F2}} 1.0 & {\cellcolor[HTML]{F2F2F2}} 0.4 & {\cellcolor[HTML]{F2F2F2}} 1.0 & {\cellcolor[HTML]{F2F2F2}} 0.6 & {\cellcolor[HTML]{F2F2F2}} 1.6 & {\cellcolor[HTML]{F2F2F2}} 0.4 & {\cellcolor[HTML]{EED3DB}} 0.9 & {\cellcolor[HTML]{F2F2F2}} 1.0 & {\cellcolor[HTML]{F2F2F2}} 1.0 & {\cellcolor[HTML]{E7ADBE}} 0.8 & {\cellcolor[HTML]{E7ADBE}} 0.8 & {\cellcolor[HTML]{E7ADBE}} 0.8 & {\cellcolor[HTML]{EED3DB}} 0.9 & {\cellcolor[HTML]{F2F2F2}} 1.0 & {\cellcolor[HTML]{F2F2F2}} 1.0 & {\cellcolor[HTML]{E7ADBE}} 0.0 & {\cellcolor[HTML]{DB6185}} 0.0 & {\cellcolor[HTML]{E7ADBE}} 0.0 & {\cellcolor[HTML]{D53D69}} -0.6 & {\cellcolor[HTML]{EED3DB}} 0.1 & {\cellcolor[HTML]{E188A2}} 0.1 & {\cellcolor[HTML]{EED3DB}} 0.2 & {\cellcolor[HTML]{F2F2F2}} 0.4 & {\cellcolor[HTML]{F2F2F2}} 1.7 & {\cellcolor[HTML]{F2F2F2}} 0.3 & {\cellcolor[HTML]{F2F2F2}} 0.4 \\

DoE(Gaussian) & {\cellcolor[HTML]{F2F2F2}} 0.2 & {\cellcolor[HTML]{DAE1EB}} 0.5 & {\cellcolor[HTML]{F2F2F2}} 0.3 & {\cellcolor[HTML]{BCCCE1}} 0.6 & {\cellcolor[HTML]{F2F2F2}} 0.4 & {\cellcolor[HTML]{F2F2F2}} 0.4 & {\cellcolor[HTML]{F2F2F2}} 0.4 & {\cellcolor[HTML]{E49AAF}} 0.7 & {\cellcolor[HTML]{F2F2F2}} 1.0 & {\cellcolor[HTML]{F2F2F2}} 1.0 & {\cellcolor[HTML]{F2F2F2}} 1.0 & {\cellcolor[HTML]{DAE1EB}} 0.4 & {\cellcolor[HTML]{E49AAF}} 0.7 & {\cellcolor[HTML]{4479BB}} 7.8 & {\cellcolor[HTML]{F2F2F2}} 1.0 & {\cellcolor[HTML]{BCCCE1}} 0.6 & {\cellcolor[HTML]{EED9DF}} 0.9 & {\cellcolor[HTML]{4479BB}} 1.3 &  & {\cellcolor[HTML]{F2F2F2}} 0.4 & {\cellcolor[HTML]{E49AAF}} 0.7 & {\cellcolor[HTML]{F2F2F2}} 1.0 & {\cellcolor[HTML]{F2F2F2}} 1.0 & {\cellcolor[HTML]{DA5C81}} 0.5 & {\cellcolor[HTML]{DF7B98}} 0.6 & {\cellcolor[HTML]{DF7B98}} 0.6 & {\cellcolor[HTML]{DF7B98}} 0.6 & {\cellcolor[HTML]{E49AAF}} 0.7 & {\cellcolor[HTML]{E9BAC8}} 0.8 & {\cellcolor[HTML]{4479BB}} 6.7 & {\cellcolor[HTML]{4479BB}} 7.9 & {\cellcolor[HTML]{4479BB}} 1.8 & {\cellcolor[HTML]{4479BB}} 2.5 & {\cellcolor[HTML]{7FA2CE}} 0.6 & {\cellcolor[HTML]{4479BB}} 4.2 & {\cellcolor[HTML]{4479BB}} 1.2 & {\cellcolor[HTML]{EED9DF}} 1.6 & {\cellcolor[HTML]{E9BAC8}} 0.1 & {\cellcolor[HTML]{F2F2F2}} 0.4 &  \\
DoE(Logistic) & {\cellcolor[HTML]{EED9DF}} 0.1 & {\cellcolor[HTML]{F2F2F2}} 0.4 & {\cellcolor[HTML]{EED9DF}} 0.2 & {\cellcolor[HTML]{F2F2F2}} 0.4 & {\cellcolor[HTML]{F2F2F2}} 0.4 & {\cellcolor[HTML]{F2F2F2}} 0.4 & {\cellcolor[HTML]{F2F2F2}} 0.4 & {\cellcolor[HTML]{DF7B98}} 0.6 & {\cellcolor[HTML]{EED9DF}} 0.9 & {\cellcolor[HTML]{EED9DF}} 0.9 & {\cellcolor[HTML]{F2F2F2}} 1.0 & {\cellcolor[HTML]{F2F2F2}} 0.3 & {\cellcolor[HTML]{E49AAF}} 0.7 & {\cellcolor[HTML]{4479BB}} 7.8 & {\cellcolor[HTML]{F2F2F2}} 1.0 & {\cellcolor[HTML]{BCCCE1}} 0.6 & {\cellcolor[HTML]{EED9DF}} 0.9 & {\cellcolor[HTML]{4479BB}} 1.3 &  & {\cellcolor[HTML]{F2F2F2}} 0.4 & {\cellcolor[HTML]{E9BAC8}} 0.8 & {\cellcolor[HTML]{DAE1EB}} 1.1 & {\cellcolor[HTML]{F2F2F2}} 1.0 & {\cellcolor[HTML]{DA5C81}} 0.5 & {\cellcolor[HTML]{DF7B98}} 0.6 & {\cellcolor[HTML]{DF7B98}} 0.6 & {\cellcolor[HTML]{E49AAF}} 0.7 & {\cellcolor[HTML]{E9BAC8}} 0.8 & {\cellcolor[HTML]{E9BAC8}} 0.8 &  & {\cellcolor[HTML]{4479BB}} 2.0 & {\cellcolor[HTML]{9DB7D7}} 0.5 & {\cellcolor[HTML]{628DC4}} 0.8 & {\cellcolor[HTML]{DAE1EB}} 0.3 & {\cellcolor[HTML]{4479BB}} 1.5 & {\cellcolor[HTML]{9DB7D7}} 0.6 & {\cellcolor[HTML]{EED9DF}} 1.6 & {\cellcolor[HTML]{E9BAC8}} 0.1 & {\cellcolor[HTML]{F2F2F2}} 0.4 & \\

\midrule
 & \begin{sideways} Asinh @ \textit{St} 1 × 1 (dof=1)\end{sideways} & 
\begin{sideways}Asinh @ \textit{St} 2 × 2 (dof=1) \end{sideways}& 
\begin{sideways}Asinh @ \textit{St} 3 × 3 (dof=2) \end{sideways}& 
\begin{sideways}Asinh @ \textit{St} 5 × 5 (dof=2)\end{sideways} &
\begin{sideways} Bimodal 1 × 1 \end{sideways}&
\begin{sideways}Bivariate \textit{Nm} 1 × 1\end{sideways} &
\begin{sideways} \textit{Hc} @ Bivariate \textit{Nm} 1 × 1\end{sideways} &
\begin{sideways} \textit{Hc} @ \textit{Mn} 25 × 25 (2-pair)\end{sideways} &
\begin{sideways} \textit{Hc} @ \textit{Mn} 3 × 3 (2-pair)\end{sideways} & 
\begin{sideways} \textit{Hc} @ \textit{Mn} 5 × 5 (2-pair) \end{sideways}& 
\begin{sideways} \textit{Mn} 2 × 2 (2-pair)\end{sideways} & 
\begin{sideways} \textit{Mn} 2 × 2 (dense)\end{sideways} &
\begin{sideways} \textit{Mn} 25 × 25 (2-pair)\end{sideways} & 
\begin{sideways} \textit{Mn} 25 × 25 (dense) \end{sideways}& 
\begin{sideways} \textit{Mn} 3 × 3 (2-pair) \end{sideways}& 
\begin{sideways} \textit{Mn} 3 × 3 (dense) \end{sideways}&
\begin{sideways} \textit{Mn} 5 × 5 (2-pair) \end{sideways}& 
\begin{sideways} \textit{Mn} 5 × 5 (dense) \end{sideways}&
\begin{sideways} \textit{Mn} 50 × 50 (dense) \end{sideways}& 
\begin{sideways} \textit{Nm} CDF @ Bivariate \textit{Nm} 1 × 1\end{sideways} &
\begin{sideways} \textit{Nm} CDF @ \textit{Mn} 25 × 25 (2-pair) \end{sideways}&
\begin{sideways} \textit{Nm} CDF @ \textit{Mn} 3 × 3 (2-pair) \end{sideways}& 
\begin{sideways} \textit{Nm} CDF @ \textit{Mn} 5 × 5 (2-pair) \end{sideways}& 
\begin{sideways} \textit{Sp} @ \textit{Mn} 25 × 25 (2-pair) \end{sideways}& 
\begin{sideways} \textit{Sp} @ \textit{Mn} 3 × 3 (2-pair)\end{sideways} & 
\begin{sideways} \textit{Sp} @ \textit{Mn} 5 × 5 (2-pair)\end{sideways} & 
\begin{sideways} \textit{Sp} @ \textit{Nm} CDF @ \textit{Mn} 25 × 25 (2-pair) \end{sideways}&
\begin{sideways} \textit{Sp} @ \textit{Nm} CDF @ \textit{Mn} 3 × 3 (2-pair) \end{sideways}& 
\begin{sideways} \textit{Sp} @ \textit{Nm} CDF @ \textit{Mn} 5 × 5 (2-pair) \end{sideways}&
\begin{sideways} \textit{St} 1 × 1 (dof=1)\end{sideways} & 
\begin{sideways} \textit{St} 2 × 2 (dof=1)\end{sideways} &
\begin{sideways} \textit{St} 2 × 2 (dof=2) \end{sideways}& 
\begin{sideways} \textit{St} 3 × 3 (dof=2)\end{sideways} & 
\begin{sideways} \textit{St} 3 × 3 (dof=3)\end{sideways} & 
\begin{sideways} \textit{St} 5 × 5 (dof=2)\end{sideways} & 
\begin{sideways} \textit{St} 5 × 5 (dof=3) \end{sideways}& 
\begin{sideways} Swiss roll 2 × 1 \end{sideways}&
\begin{sideways} Uniform 1 × 1 (additive noise=0.1) \end{sideways}&
\begin{sideways} Uniform 1 × 1 (additive noise=0.75) \end{sideways}& 
\begin{sideways} Wiggly @ Bivariate \textit{Nm} 1 × 1 \end{sideways} \\

\bottomrule
\end{tabular}}

\caption{Mean MI estimates over 10 seeds using N = 10k test samples against ground truth (GT). Color indicates relative
negative (red) and positive bias (blue). All methods were trained with 100k samples. List of abbreviations ( \textit{Mn}: Multinormal,  \textit{St}: Student-t, \textit{Nm}: Normal, \textit{Hc}: Half-cube, \textit{Sp}: Spiral)
}
\label{tab:benchmark}

\end{table}

\vspace*{-1cm}
\section{Conclusion}\label{sec:conclusion}
The estimation of \gls{MI} stands as a fundamental goal in many areas of machine learning, as it enables understanding the relationships within data, driving representation learning, and evaluating generative models. Over the years, various methodologies have emerged to tackle the difficult task of \gls{MI} estimation, addressing challenges posed by high-dimensional, real-world data. Our work introduced a novel method, \gls{MINDE}, which provides a unique perspective on \gls{MI} estimation by leveraging the theory of diffusion-based generative models. We expanded the classical toolkit for information-theoretic analysis, and showed how to compute the \gls{KL} divergence and entropy of random variables using the score of data distributions. We defined several variants of \gls{MINDE}, which we have extensively tested according to a recent, comprehensive benchmark that simulates real-world challenges, including sparsity, long-tailed distributions, invariance to transformations. Our results indicated that our methods outperform state-of-the-art alternatives, especially on the most challenging tests. Additionally, \gls{MINDE} variants successfully passed self-consistency tests, validating the robustness and reliability of our proposed methodology.

Our research opens up exciting avenues for future exploration. One compelling direction is the application of \gls{MINDE} to large-scale multi-modal datasets. The conditional version of our approach enables harnessing the extensive repository of existing pre-trained diffusion models. For instance, it could find valuable application in the estimation of \gls{MI} for text-conditional image generation. Conversely, our joint modeling approach offers a straightforward path to scaling \gls{MI} estimation to more than two variables. A scalable approach to \gls{MI} estimation is particularly valuable when dealing with complex systems involving multiple interacting variables, eliminating the need to specify a hierarchy among them.

\section*{Acknowledgements}
GF and PM gratefully acknowledges support from Huawei Paris and the European Commission (ADROIT6G
Grant agreement ID: 101095363).


\bibliography{biblio.bib}

\begin{thebibliography}{55}
\providecommand{\natexlab}[1]{#1}
\providecommand{\url}[1]{\texttt{#1}}
\expandafter\ifx\csname urlstyle\endcsname\relax
  \providecommand{\doi}[1]{doi: #1}\else
  \providecommand{\doi}{doi: \begingroup \urlstyle{rm}\Url}\fi

\bibitem[Alemi \& Fischer(2018)Alemi and Fischer]{alemi2019gilbo}
Alexander~A Alemi and Ian Fischer.
\newblock Gilbo: One metric to measure them all.
\newblock \emph{Advances in Neural Information Processing Systems}, 31, 2018.

\bibitem[Alemi et~al.(2016)Alemi, Fischer, Dillon, and Murphy]{alemi2019deep}
Alexander~A Alemi, Ian Fischer, Joshua~V Dillon, and Kevin Murphy.
\newblock Deep variational information bottleneck.
\newblock In \emph{International Conference on Learning Representations}, 2016.

\bibitem[Anderson(1982)]{anderson1982reverse}
Brian D.~O. Anderson.
\newblock Reverse-time diffusion equation models.
\newblock \emph{Stochastic Processes and their Applications}, 12\penalty0
  (3):\penalty0 313--326, 1982.

\bibitem[Balaji et~al.(2022)Balaji, Nah, Huang, Vahdat, Song, Kreis, Aittala,
  Aila, Laine, Catanzaro, et~al.]{balaji2022ediffi}
Yogesh Balaji, Seungjun Nah, Xun Huang, Arash Vahdat, Jiaming Song, Karsten
  Kreis, Miika Aittala, Timo Aila, Samuli Laine, Bryan Catanzaro, et~al.
\newblock ediffi: Text-to-image diffusion models with an ensemble of expert
  denoisers.
\newblock \emph{arXiv preprint arXiv:2211.01324}, 2022.

\bibitem[Bao et~al.(2023)Bao, Nie, Xue, Li, Pu, Wang, Yue, Cao, Su, and
  Zhu]{bao2023transformer}
Fan Bao, Shen Nie, Kaiwen Xue, Chongxuan Li, Shi Pu, Yaole Wang, Gang Yue, Yue
  Cao, Hang Su, and Jun Zhu.
\newblock One transformer fits all distributions in multi-modal diffusion at
  scale.
\newblock \emph{arXiv preprint arXiv:2303.06555}, 2023.

\bibitem[Barber \& Agakov(2004)Barber and Agakov]{barber2004algorithm}
David Barber and Felix Agakov.
\newblock The im algorithm: a variational approach to information maximization.
\newblock \emph{Advances in neural information processing systems}, 16\penalty0
  (320):\penalty0 201, 2004.

\bibitem[Belghazi et~al.(2018)Belghazi, Baratin, Rajeshwar, Ozair, Bengio,
  Courville, and Hjelm]{belghazi2018mine}
Mohamed~Ishmael Belghazi, Aristide Baratin, Sai Rajeshwar, Sherjil Ozair,
  Yoshua Bengio, Aaron Courville, and Devon Hjelm.
\newblock Mutual information neural estimation.
\newblock In \emph{Proceedings of the 35th International Conference on Machine
  Learning}, 2018.

\bibitem[Bell \& Sejnowski(1995)Bell and Sejnowski]{bell1995information}
Anthony~J Bell and Terrence~J Sejnowski.
\newblock An information-maximization approach to blind separation and blind
  deconvolution.
\newblock \emph{Neural computation}, 7\penalty0 (6):\penalty0 1129--1159, 1995.

\bibitem[Bounoua et~al.(2023)Bounoua, Franzese, and
  Michiardi]{bounoua2023multimodal}
Mustapha Bounoua, Giulio Franzese, and Pietro Michiardi.
\newblock Multi-modal latent diffusion.
\newblock \emph{arXiv preprint arXiv:2306.04445}, 2023.

\bibitem[Brekelmans et~al.(2022)Brekelmans, Huang, Ghassemi, Steeg, Grosse, and
  Makhzani]{brekelmans2023improving}
Rob Brekelmans, Sicong Huang, Marzyeh Ghassemi, Greg~Ver Steeg, Roger~Baker
  Grosse, and Alireza Makhzani.
\newblock Improving mutual information estimation with annealed and
  energy-based bounds.
\newblock In \emph{International Conference on Learning Representations}, 2022.

\bibitem[Chen et~al.(2022)Chen, Chewi, Li, Li, Salim, and
  Zhang]{chen2022sampling}
Sitan Chen, Sinho Chewi, Jerry Li, Yuanzhi Li, Adil Salim, and Anru Zhang.
\newblock Sampling is as easy as learning the score: theory for diffusion
  models with minimal data assumptions.
\newblock In \emph{International Conference on Learning Representations}, 2022.

\bibitem[Chen et~al.(2016)Chen, Duan, Houthooft, Schulman, Sutskever, and
  Abbeel]{chen2016infogan}
Xi~Chen, Yan Duan, Rein Houthooft, John Schulman, Ilya Sutskever, and Pieter
  Abbeel.
\newblock Infogan: Interpretable representation learning by information
  maximizing generative adversarial nets.
\newblock \emph{Advances in neural information processing systems}, 29, 2016.

\bibitem[Collet \& Malrieu(2008)Collet and Malrieu]{collet2008logarithmic}
Jean-Fran{\c{c}}ois Collet and Florent Malrieu.
\newblock Logarithmic sobolev inequalities for inhomogeneous markov semigroups.
\newblock \emph{ESAIM: Probability and Statistics}, 12:\penalty0 492--504,
  2008.

\bibitem[Czy{\.z} et~al.(2023)Czy{\.z}, Grabowski, Vogt, Beerenwinkel, and
  Marx]{czyz2023beyond}
Pawe{\l} Czy{\.z}, Frederic Grabowski, Julia~E Vogt, Niko Beerenwinkel, and
  Alexander Marx.
\newblock Beyond normal: On the evaluation of mutual information estimators.
\newblock \emph{Advances in Neural Information Processing Systems}, 2023.

\bibitem[De~Bortoli(2022)]{de2022convergence}
Valentin De~Bortoli.
\newblock Convergence of denoising diffusion models under the manifold
  hypothesis.
\newblock \emph{Transactions on Machine Learning Research}, 2022.

\bibitem[Federici et~al.(2023)Federici, Ruhe, and
  Forr{\'e}]{federici2023effectiveness}
Marco Federici, David Ruhe, and Patrick Forr{\'e}.
\newblock On the effectiveness of hybrid mutual information estimation.
\newblock \emph{arXiv preprint arXiv:2306.00608}, 2023.

\bibitem[Franzese et~al.(2023)Franzese, Rossi, Yang, Finamore, Rossi,
  Filippone, and Michiardi]{franzese2022}
Giulio Franzese, Simone Rossi, Lixuan Yang, Alessandro Finamore, Dario Rossi,
  Maurizio Filippone, and Pietro Michiardi.
\newblock How much is enough? a study on diffusion times in score-based
  generative models.
\newblock \emph{Entropy}, 2023.

\bibitem[Gao et~al.(2015)Gao, Ver~Steeg, and Galstyan]{gao2015efficient}
Shuyang Gao, Greg Ver~Steeg, and Aram Galstyan.
\newblock Efficient estimation of mutual information for strongly dependent
  variables.
\newblock In \emph{Artificial intelligence and statistics}, pp.\  277--286.
  PMLR, 2015.

\bibitem[Hjelm et~al.(2019)Hjelm, Fedorov, Lavoie-Marchildon, Grewal, Bachman,
  Trischler, and Bengio]{hjelm2019learning}
R~Devon Hjelm, Alex Fedorov, Samuel Lavoie-Marchildon, Karan Grewal, Phil
  Bachman, Adam Trischler, and Yoshua Bengio.
\newblock Learning deep representations by mutual information estimation and
  maximization.
\newblock In \emph{International Conference on Learning Representations}, 2019.

\bibitem[Ho \& Salimans(2021)Ho and Salimans]{ho2021classifier}
Jonathan Ho and Tim Salimans.
\newblock Classifier-free diffusion guidance.
\newblock In \emph{NeurIPS 2021 Workshop on Deep Generative Models and
  Downstream Applications}, 2021.

\bibitem[Ho et~al.(2020)Ho, Jain, and Abbeel]{ho2020}
Jonathan Ho, Ajay Jain, and Pieter Abbeel.
\newblock Denoising diffusion probabilistic models.
\newblock In H.~Larochelle, M.~Ranzato, R.~Hadsell, M.F. Balcan, and H.~Lin
  (eds.), \emph{Advances in Neural Information Processing Systems}, volume~33,
  pp.\  6840--6851. Curran Associates, Inc., 2020.

\bibitem[Huang et~al.(2021)Huang, Lim, and Courville]{huang2021variational}
Chin-Wei Huang, Jae~Hyun Lim, and Aaron~C Courville.
\newblock A variational perspective on diffusion-based generative models and
  score matching.
\newblock \emph{Advances in Neural Information Processing Systems},
  34:\penalty0 22863--22876, 2021.

\bibitem[Huang et~al.(2020)Huang, Makhzani, Cao, and
  Grosse]{huang2020evaluating}
Sicong Huang, Alireza Makhzani, Yanshuai Cao, and Roger Grosse.
\newblock Evaluating lossy compression rates of deep generative models.
\newblock In \emph{International Conference on Machine Learning}. PMLR, 2020.

\bibitem[Kingma \& Ba(2015)Kingma and Ba]{kingma2014adam}
Diederik Kingma and Jimmy Ba.
\newblock Adam: A method for stochastic optimization.
\newblock In \emph{International Conference on Learning Representations
  (ICLR)}, 2015.

\bibitem[Kingma et~al.(2021)Kingma, Salimans, Poole, and Ho]{kingma2021}
Diederik~P Kingma, Tim Salimans, Ben Poole, and Jonathan Ho.
\newblock Variational diffusion models.
\newblock In A.~Beygelzimer, Y.~Dauphin, P.~Liang, and J.~Wortman Vaughan
  (eds.), \emph{Advances in Neural Information Processing Systems}, 2021.

\bibitem[Kong et~al.(2022)Kong, Brekelmans, and Ver~Steeg]{kong2023information}
Xianghao Kong, Rob Brekelmans, and Greg Ver~Steeg.
\newblock Information-theoretic diffusion.
\newblock In \emph{International Conference on Learning Representations}, 2022.

\bibitem[Kraskov et~al.(2004)Kraskov, St{\"o}gbauer, and
  Grassberger]{kraskov2004estimating}
Alexander Kraskov, Harald St{\"o}gbauer, and Peter Grassberger.
\newblock Estimating mutual information.
\newblock \emph{Physical review E}, 69\penalty0 (6):\penalty0 066138, 2004.

\bibitem[Lee et~al.(2022)Lee, Lu, and Tan]{lee2022convergence}
Holden Lee, Jianfeng Lu, and Yixin Tan.
\newblock Convergence for score-based generative modeling with polynomial
  complexity.
\newblock \emph{Advances in Neural Information Processing Systems},
  35:\penalty0 22870--22882, 2022.

\bibitem[L{\'e}onard(2014)]{leonard2014some}
Christian L{\'e}onard.
\newblock Some properties of path measures.
\newblock \emph{S{\'e}minaire de Probabilit{\'e}s XLVI}, pp.\  207--230, 2014.

\bibitem[Letizia \& Tonello(2022)Letizia and Tonello]{letizia2022copula}
Nunzio~A Letizia and Andrea~M Tonello.
\newblock Copula density neural estimation.
\newblock \emph{arXiv preprint arXiv:2211.15353}, 2022.

\bibitem[Letizia et~al.(2023)Letizia, Novello, and
  Tonello]{letizia2023variational}
Nunzio~A Letizia, Nicola Novello, and Andrea~M Tonello.
\newblock Variational $ f $-divergence and derangements for discriminative
  mutual information estimation.
\newblock \emph{arXiv preprint arXiv:2305.20025}, 2023.

\bibitem[Loaiza-Ganem et~al.(2022)Loaiza-Ganem, Ross, Cresswell, and
  Caterini]{loaiza2022diagnosing}
Gabriel Loaiza-Ganem, Brendan~Leigh Ross, Jesse~C Cresswell, and Anthony~L
  Caterini.
\newblock Diagnosing and fixing manifold overfitting in deep generative models.
\newblock \emph{Transactions on Machine Learning Research}, 2022.

\bibitem[MacKay(2003)]{mackay2003information}
David~JC MacKay.
\newblock \emph{Information theory, inference and learning algorithms}.
\newblock Cambridge university press, 2003.

\bibitem[McAllester \& Stratos(2020)McAllester and
  Stratos]{mcallester2020formal}
David McAllester and Karl Stratos.
\newblock Formal limitations on the measurement of mutual information.
\newblock In \emph{International Conference on Artificial Intelligence and
  Statistics}, 2020.

\bibitem[Moon et~al.(1995)Moon, Rajagopalan, and Lall]{moon1995estimation}
Young-Il Moon, Balaji Rajagopalan, and Upmanu Lall.
\newblock Estimation of mutual information using kernel density estimators.
\newblock \emph{Physical Review E}, 52\penalty0 (3):\penalty0 2318, 1995.

\bibitem[Nguyen et~al.(2007)Nguyen, Wainwright, and Jordan]{nguyen2007neurips}
XuanLong Nguyen, Martin~J Wainwright, and Michael Jordan.
\newblock Estimating divergence functionals and the likelihood ratio by
  penalized convex risk minimization.
\newblock In \emph{Advances in Neural Information Processing Systems}, 2007.

\bibitem[Nowozin et~al.(2016)Nowozin, Cseke, and Tomioka]{nowozin2016neurips}
Sebastian Nowozin, Botond Cseke, and Ryota Tomioka.
\newblock F-gan: Training generative neural samplers using variational
  divergence minimization.
\newblock In \emph{Proceedings of the 30th International Conference on Neural
  Information Processing Systems}, NIPS'16, pp.\  271–279, Red Hook, NY, USA,
  2016. Curran Associates Inc.
\newblock ISBN 9781510838819.

\bibitem[{\O}ksendal(2003)]{oksendal2003stochastic}
Bernt {\O}ksendal.
\newblock \emph{Stochastic differential equations}.
\newblock Springer, 2003.

\bibitem[Oord et~al.(2018)Oord, Li, and Vinyals]{oord2019representation}
Aaron van~den Oord, Yazhe Li, and Oriol Vinyals.
\newblock Representation learning with contrastive predictive coding.
\newblock \emph{Advances in neural information processing systems}, 2018.

\bibitem[Paninski(2003)]{paninski2003estimation}
Liam Paninski.
\newblock Estimation of entropy and mutual information.
\newblock \emph{Neural computation}, 15\penalty0 (6):\penalty0 1191--1253,
  2003.

\bibitem[Papamakarios et~al.(2017)Papamakarios, Pavlakou, and
  Murray]{papamakarios2017masked}
George Papamakarios, Theo Pavlakou, and Iain Murray.
\newblock Masked autoregressive flow for density estimation.
\newblock \emph{Advances in neural information processing systems}, 30, 2017.

\bibitem[Pizer et~al.(1987)Pizer, Amburn, Austin, Cromartie, Geselowitz, Greer,
  ter Haar~Romeny, Zimmerman, and Zuiderveld]{pizer1987adaptive}
Stephen~M Pizer, E~Philip Amburn, John~D Austin, Robert Cromartie, Ari
  Geselowitz, Trey Greer, Bart ter Haar~Romeny, John~B Zimmerman, and Karel
  Zuiderveld.
\newblock Adaptive histogram equalization and its variations.
\newblock \emph{Computer vision, graphics, and image processing}, 39\penalty0
  (3):\penalty0 355--368, 1987.

\bibitem[Poole et~al.(2019)Poole, Ozair, Van Den~Oord, Alemi, and
  Tucker]{poole2019variational}
Ben Poole, Sherjil Ozair, Aaron Van Den~Oord, Alex Alemi, and George Tucker.
\newblock On variational bounds of mutual information.
\newblock In \emph{International Conference on Machine Learning}, 2019.

\bibitem[Rainforth et~al.(2018)Rainforth, Cornish, Yang, Warrington, and
  Wood]{rainforth2018nesting}
Tom Rainforth, Rob Cornish, Hongseok Yang, Andrew Warrington, and Frank Wood.
\newblock On nesting monte carlo estimators.
\newblock In \emph{International Conference on Machine Learning}, pp.\
  4267--4276. PMLR, 2018.

\bibitem[Ramesh et~al.(2022)Ramesh, Dhariwal, Nichol, Chu, and
  Chen]{ramesh2022}
Aditya Ramesh, Prafulla Dhariwal, Alex Nichol, Casey Chu, and Mark Chen.
\newblock Hierarchical text-conditional image generation with clip latents,
  2022.
\newblock URL \url{https://arxiv.org/abs/2204.06125}.

\bibitem[Rhodes et~al.(2020)Rhodes, Xu, and Gutmann]{rhodes2020telescoping}
Benjamin Rhodes, Kai Xu, and Michael~U Gutmann.
\newblock Telescoping density-ratio estimation.
\newblock \emph{Advances in neural information processing systems}, 2020.

\bibitem[Rombach et~al.(2022)Rombach, Blattmann, Lorenz, Esser, and
  Ommer]{rombach2022}
Robin Rombach, Andreas Blattmann, Dominik Lorenz, Patrick Esser, and Bj\"orn
  Ommer.
\newblock High-resolution image synthesis with latent diffusion models.
\newblock In \emph{Proceedings of the IEEE/CVF Conference on Computer Vision
  and Pattern Recognition (CVPR)}, pp.\  10684--10695, June 2022.

\bibitem[Saharia et~al.(2022)Saharia, Chan, Saxena, Li, Whang, Denton,
  Ghasemipour, Gontijo-Lopes, Ayan, Salimans, Ho, Fleet, and
  Norouzi]{saharia2022}
Chitwan Saharia, William Chan, Saurabh Saxena, Lala Li, Jay Whang, Emily
  Denton, Seyed Kamyar~Seyed Ghasemipour, Raphael Gontijo-Lopes, Burcu~Karagol
  Ayan, Tim Salimans, Jonathan Ho, David~J. Fleet, and Mohammad Norouzi.
\newblock Photorealistic text-to-image diffusion models with deep language
  understanding.
\newblock In Alice~H. Oh, Alekh Agarwal, Danielle Belgrave, and Kyunghyun Cho
  (eds.), \emph{Advances in Neural Information Processing Systems}, 2022.

\bibitem[Schuhmann et~al.(2022)Schuhmann, Beaumont, Vencu, Gordon, Wightman,
  Cherti, Coombes, Katta, Mullis, Wortsman, Schramowski, Kundurthy, Crowson,
  Schmidt, Kaczmarczyk, and Jitsev]{schuhmann2022laion5b}
Christoph Schuhmann, Romain Beaumont, Richard Vencu, Cade Gordon, Ross
  Wightman, Mehdi Cherti, Theo Coombes, Aarush Katta, Clayton Mullis, Mitchell
  Wortsman, Patrick Schramowski, Srivatsa Kundurthy, Katherine Crowson, Ludwig
  Schmidt, Robert Kaczmarczyk, and Jenia Jitsev.
\newblock Laion-5b: An open large-scale dataset for training next generation
  image-text models, 2022.

\bibitem[Shannon(1948)]{shannon1948mathematical}
Claude~Elwood Shannon.
\newblock A mathematical theory of communication.
\newblock \emph{The Bell system technical journal}, 27\penalty0 (3):\penalty0
  379--423, 1948.

\bibitem[Song \& Ermon(2019)Song and Ermon]{song2019understanding}
Jiaming Song and Stefano Ermon.
\newblock Understanding the limitations of variational mutual information
  estimators.
\newblock In \emph{International Conference on Learning Representations}, 2019.

\bibitem[Song et~al.(2021)Song, Sohl-Dickstein, Kingma, Kumar, Ermon, and
  Poole]{song2021a}
Yang Song, Jascha Sohl-Dickstein, Diederik~P Kingma, Abhishek Kumar, Stefano
  Ermon, and Ben Poole.
\newblock Score-based generative modeling through stochastic differential
  equations.
\newblock In \emph{International Conference on Learning Representations}, 2021.

\bibitem[Stratos(2019)]{stratos2018mutual}
Karl Stratos.
\newblock Mutual information maximization for simple and accurate
  part-of-speech induction.
\newblock In \emph{Proceedings of the 2019 Conference of the North American
  Chapter of the Association for Computational Linguistics: Human Language
  Technologies, Volume 1 (Long and Short Papers)}, 2019.

\bibitem[Wunder et~al.(2021)Wunder, Gro{\ss}, Fritschek, and
  Schaefer]{wunder2021reverse}
Gerhard Wunder, Benedikt Gro{\ss}, Rick Fritschek, and Rafael~F Schaefer.
\newblock A reverse jensen inequality result with application to mutual
  information estimation.
\newblock In \emph{2021 IEEE Information Theory Workshop (ITW)}, 2021.

\bibitem[Zhao et~al.(2018)Zhao, Song, and Ermon]{zhao2018information}
Shengjia Zhao, Jiaming Song, and Stefano Ermon.
\newblock A lagrangian perspective on latent variable generative models.
\newblock In \emph{Proc. 34th Conference on Uncertainty in Artificial
  Intelligence}, 2018.

\end{thebibliography}
\bibliographystyle{iclr2024_conference}


\setcounter{section}{0}
\renewcommand{\thesection}{\Alph{section}}
\renewcommand{\theHsection}{appendixsection.\Alph{section}}

\section*{MINDE: Mutual Information Neural Diffusion Estimation --- Supplementary material}

\section{Proofs of \Cref{sec:kl_deriv}}\label{sec:proof_kl_deriv}
\emph{Proof of Auto-encoder invariance of \gls{KL}}. Whenever we can find encoder and decoder functions $\phi,\psi$ respectively such that $\phi(\psi(x))=x,\mu^A-$ almost surely and $\phi(\psi(x))=x,\mu^B-$ almost surely, the Kullback-Leibler divergence can be computed in the \textit{latent} space obtained by the encoder $\psi$:

\begin{flalign}
    &\KL{\mu^A}{\mu^B}=\int_{\mathcal{M}}\log\frac{\dd \mu^A}{\dd \mu^B}\dd \mu^A=\nonumber\\
    &\int_{\mathcal{M}}\log\left(\frac{\dd \mu^A}{\dd \mu^B}\circ\phi\circ\psi\right)\dd \mu^A=
    \int_{\psi(\mathcal{M})}\log\left(\frac{\dd \mu^A}{\dd \mu^B}\circ\phi\right)\dd ({\mu}^A\circ \psi^{-1})=\nonumber\\&\int_{\psi(\mathcal{M})}\log\left(\frac{\dd \mu^A}{\dd \mu^B}\circ\psi^{(-1)}\right)\dd ({\mu}^A\circ \psi^{-1})=
    \KL{\tilde\mu^A}{\tilde\mu^B}.
\end{flalign}

\emph{Proof of \Cref{error_score}}. To prove such claim, it is sufficient to start from the r.h.s. of \Cref{eq:estimator}, substitute to the parametric scores their definition with the errors $\epsilon^{\mu^A}_t(x)= \tilde{s}^{\mu^A}_t(x)-s^{\mu^A}_t(x)$, and expand the square:
\begin{flalign*}
   & \int\limits_0^T\frac{g^2_{t}}{2} \E_{\nu^{\mu^A}_t}\left[\norm{\tilde{s}^{\mu^A}_t(X_t)-\tilde{s}^{\mu^B}_t(X_t) }^2\right]\dd t=\\
   &\int\limits_0^T\frac{g^2_{t}}{2} \E_{\nu^{\mu^A}_t}\left[\norm{{s}^{\mu^A}_t(X_t)+\epsilon^{\mu^A}_t(x)-{s}^{\mu^B}_t(X_t) -\epsilon^{\mu^B}_t(x)}^2\right]\dd t=\\
   &\int\limits_0^T\frac{g^2_{t}}{2} \E_{\nu^{\mu^A}_t}\left[\norm{{s}^{\mu^A}_t(X_t)-{s}^{\mu^B}_t(X_t) }^2\right]\dd t+\\&\int\limits_0^T\frac{g^2_{t}}{2}\E_{\nu^{\mu^A}_t}\left[\norm{\epsilon^{\mu^A}_t(X_t)-\epsilon^{\mu^B}_t(X_t)}^2+2\langle {s}^{\mu^A}_t(X_t)-{s}^{\mu^B}_t(X_t) +,\epsilon^{\mu^A}_t(X_t)-\epsilon^{\mu^B}_t(X_t)\rangle\right]\dd t,
\end{flalign*}
from which the definition of $d$ holds.

\section{Proof of \Cref{eq:mi_3inner}}\label{proof_eq:mi_3inner}

We start with the approximation of \Cref{eq:mi_3}:
\begin{equation}
    \textsc{I}(A,B)\simeq-e(\mu^{C},\GaussM{\sigma})+\int e(\mu^{A_y},\GaussM{\sigma})\dd\mu^{B}(y)+\int e(\mu^{B_x},\GaussM{\sigma})\dd\mu^{A}(x). 
\end{equation}
Since the approximation is valid for any $\sigma$, we select the limit of $\sigma\rightarrow\infty$, where the reference score $\chi^{-1}_tx$ converges to zero, and can thus be neglected from the estimators integral (for example, $e(\mu^A,\GaussM{\infty})\simeq \int\limits_0^T\frac{g^2_{t}}{2} \E_{\nu^{\mu^A}_t}\left[\norm{\tilde{s}^{\mu^A}_t(X_t) }^2\right]\dd t$). This allows to obtain:
\begin{flalign*}
   \textsc{I}(A,B) &\simeq -\int\limits_0^T\frac{g^2_{t}}{2} \int \dd\nu^{\mu^C}_t([x,y])\norm{\tilde{s}^{\mu^C}_t([x,y])}^2\dd t+\\
   &\int \left(\int\limits_0^T\frac{g^2_{t}}{2} \int \dd\nu^{\mu^{A_y}}_t(x)\norm{\tilde{s}^{\mu^{A_y}}_t(x)}^2\dd t\right)\dd\mu^B(y)+\\
   & \int \left(\int\limits_0^T\frac{g^2_{t}}{2} \int \dd\nu^{\mu^{B_x}}_t(y)\norm{\tilde{s}^{\mu^{B_x}}_t(y)}^2\dd t\right)\dd\mu^A(x).
\end{flalign*}
As a further step in the derivation of our approximation, we consider the estimated scores to be sufficiently good, such that we substitute the parametric with the true scores. In such case, the following holds:
\begin{flalign*}
    &\textsc{I}(A,B) \simeq \\
    &\int\limits_0^T\frac{g^2_{t}}{2} \int \dd\mu^C([x_0,y_0])\dd\nu_t^{\delta_{[x_0,y_0]}}([x,y])\left(-\norm{{s}^{\mu^C}_t([x,y])}^2+\norm{{s}^{\mu^{A_{y_0}}}_t(x)}^2+\norm{{s}^{\mu^{B_{x_0}}}_t(y)}^2\right)\dd t= \\
    &\int\limits_0^T\frac{g^2_{t}}{2} \int \dd\mu^C([x_0,y_0])\dd\nu_t^{\delta_{[x_0,y_0]}}([x,y])\left(-\norm{{s}^{\mu^C}_t([x,y])}^2+\norm{[{s}^{\mu^{A_{y_0}}}_t(x),{s}^{\mu^{B_{x_0}}}_t(y)]}^2\right)\dd t=\\
    &\int\limits_0^T\frac{g^2_{t}}{2} \int \dd\mu^C([x_0,y_0])\dd\nu_t^{\delta_{[x_0,y_0]}}([x,y])\left(-2\norm{{s}^{\mu^C}_t([x,y])}^2+\norm{{s}^{\mu^C}_t([x,y])-[{s}^{\mu^{A_{y_0}}}_t(x),{s}^{\mu^{B_{x_0}}}_t(y)]}^2+\right.\\& \left.2 \left\langle{s}^{\mu^C}_t([x,y]),[{s}^{\mu^{A_{y_0}}}_t(x),{s}^{\mu^{B_{x_0}}}_t(y)]\right\rangle\right)\dd t.
\end{flalign*}

Recognizing that the term $\norm{{s}^{\mu^C}_t([x,y])-[{s}^{\mu^{A_{y_0}}}_t(x),{s}^{\mu^{B_{x_0}}}_t(y)]}^2$, averaged over the measures, is just \Cref{eq:mi_3inner} in disguise, what remain to be assessed is the following:
\begin{flalign}\label{eq:pr0}
  \int\limits_0^T\frac{g^2_{t}}{2} &\int \dd\mu^C([x_0,y_0])\dd\nu_t^{\delta_{[x_0,y_0]}}([x,y]) \nonumber\\
  &\left(-2\norm{{s}^{\mu^C}_t([x,y])}^2+2 \left\langle{s}^{\mu^C}_t([x,y]),[{s}^{\mu^{A_{y_0}}}_t(x),{s}^{\mu^{B_{x_0}}}_t(y)]\right\rangle\right)\dd t=0.
\end{flalign}

In particular, we focus on the term:
\begin{flalign} \label{refprove}
     & \int\limits_0^T\frac{g^2_{t}}{2} \int \dd\mu^C([x_0,y_0])\dd\nu_t^{\delta_{[x_0,y_0]}}([x,y]) \left\langle{s}^{\mu^C}_t([x,y]),[{s}^{\mu^{A_{y_0}}}_t(x),{s}^{\mu^{B_{x_0}}}_t(y)]\right\rangle\dd t=\nonumber\\
     & \int\limits_0^T\frac{g^2_{t}}{2} \int_{x,y} \left\langle{s}^{\mu^C}_t([x,y]),\right.\nonumber\\&\left.\left[\int_{x_0,y_0}\dd\mu^C([x_0,y_0])\dd\nu_t^{\delta_{[x_0,y_0]}}([x,y]){s}^{\mu^{A_{y_0}}}_t(x),
     \int_{x_0,y_0}\dd\mu^C([x_0,y_0])\dd\nu_t^{\delta_{[x_0,y_0]}}([x,y]){s}^{\mu^{B_{x_0}}}_t(y)\right]\right\rangle\dd t.
\end{flalign}

Since $\dd\nu_t^{\delta_{[x_0,y_0]}}([x,y])=\dd\nu_t^{\delta_{x_0}}(x)\dd\nu_t^{\delta_{y_0}}(y)$ and $\dd\mu^C([x_0,y_0])=\dd\mu^{A_{y_0}}(x_0)\dd\mu^B(y_0)$, then $\int_{x_0}\dd\mu^C([x_0,y_0])\dd\nu_t^{\delta_{[x_0,y_0]}}([x,y])=\dd\nu^{\mu^{A_{y_0}}}_t(x)\dd\nu_t^{\delta_{y_0}}(y)\dd\mu^B(y_0)$. Consequently:
\begin{flalign*}
  &\int_{x_0,y_0}\dd\mu^C([x_0,y_0])\dd\nu_t^{\delta_{[x_0,y_0]}}([x,y]){s}^{\mu^{A_{y_0}}}_t(x)=\int_{y_0}\dd\nu^{\mu^{A_{y_0}}}_t(x)\dd\nu_t^{\delta_{y_0}}(y)\dd\mu^B(y_0){s}^{\mu^{A_{y_0}}}_t(x)=\\&
  \int_{y_0}\dd\nu^{\mu^{A_{y_0}}}_t(x)\dd\nu_t^{\delta_{y_0}}(y)\dd\mu^B(y_0)\nabla\log\left(\bar\nu^{\mu^{A_{y_0}}}_t(x)\right)=
  \int_{y_0}\dd\nu^{\mu^{A_{y_0}}}_t(x)\dd\nu_t^{\delta_{y_0}}(y)\dd\mu^B(y_0)\frac{\nabla\bar\nu^{\mu^{A_{y_0}}}_t(x)}{\bar\nu^{\mu^{A_{y_0}}}_t(x)}=\\
  &\dd x\int_{y_0}\dd\nu_t^{\delta_{y_0}}(y)\dd\mu^B(y_0)\nabla\bar\nu^{\mu^{A_{y_0}}}_t(x)=
  \dd x\nabla\left(\int_{y_0}\dd\nu_t^{\delta_{y_0}}(y)\dd\mu^B(y_0)\bar\nu^{\mu^{A_{y_0}}}_t(x)\right)=\\
  &\dd x\dd\nu^{\mu^{B}}_t(y)\nabla\left(\int_{y_0}\dd\mu^{B\g Y_t=y}(y_0)\bar\nu^{\mu^{A_{y_0}}}_t(x)\right)=\dd x\dd\nu^{\mu^{B}}_t(y)\nabla\left(\bar\nu^{\mu^{A\g Y_t=y}}_t(x)\right),
\end{flalign*}
where in the last line we introduced: $\mu^{B\g Y_t=y}(y_0)$, the measure of the random variable $B$ conditionally on the fact that the diffused variable $B$ after a time $t$ is equal to $y$ and $\nu^{\mu^{A\g Y_t=y}}$, the conditional measure of the diffused variable $A$ at time $t$, conditionally on the diffused variable $B$ after a time $t$ equal to $y$. Finally
\begin{flalign*}
 &\dd x\dd\nu^{\mu^{B}}_t(y)\nabla\left(\bar\nu^{\mu^{A\g Y_t=y}}_t(x)\right)=\bar\nu^{\mu^{A\g Y_t=y}}_t(x)\dd x\dd\nu^{\mu^{B}}_t(y)\frac{\nabla\left(\bar\nu^{\mu^{A\g Y_t=y}}_t(x)\right)}{\bar\nu^{\mu^{A\g Y_t=y}}_t(x)}=\dd \nu^{\mu^{C}}_t([x,y])s^{\mu^{A\g Y_t=y}}_t(x).
\end{flalign*}
Along the same lines, we can prove the equality $\int_{x_0,y_0}\dd\mu^C([x_0,y_0])\dd\nu_t^{\delta_{[x_0,y_0]}}([x,y]){s}^{\mu^{B_{x_0}}}_t(y)=\dd \nu^{\mu^{C}}_t([x,y])s^{\mu^{B\g X_t=x}}_t(y)$. Then, restarting from \Cref{refprove}
we have:
\begin{flalign*}
    &\int\limits_0^T\frac{g^2_{t}}{2} \int_{x,y} \left\langle{s}^{\mu^C}_t([x,y]),\right.\nonumber\\&\left.\left[\int_{x_0,y_0}\dd\mu^C([x_0,y_0])\dd\nu_t^{\delta_{[x_0,y_0]}}([x,y]){s}^{\mu^{A_{y_0}}}_t(x),\int_{x_0,y_0}\dd\mu^C([x_0,y_0])\dd\nu_t^{\delta_{[x_0,y_0]}}([x,y]){s}^{\mu^{B_{x_0}}}_t(y)\right]\right\rangle\dd t=\\&
    \int\limits_0^T\frac{g^2_{t}}{2} \int_{x,y} \left\langle{s}^{\mu^C}_t([x,y]),\left[\dd \nu^{\mu^{C}}_t([x,y])s^{\mu^{A\g Y_t=y}}_t(x),\dd \nu^{\mu^{C}}_t([x,y])s^{\mu^{B\g X_t=x}}_t(y)\right]\right\rangle\dd t=\\&\int\limits_0^T\frac{g^2_{t}}{2} \int_{x,y} \dd \nu^{\mu^{C}}_t([x,y])\left\langle{s}^{\mu^C}_t([x,y]),[s^{\mu^{A\g Y_t=y}}_t(x),s^{\mu^{B\g X_t=x}}_t(y)]\right\rangle\dd t=\int\limits_0^T\frac{g^2_{t}}{2} \int_{x,y} \dd \nu^{\mu^{C}}_t([x,y])\norm{{s}^{\mu^C}_t([x,y])}^2\dd t,
\end{flalign*}
which finally allows to prove \Cref{eq:pr0} and claim validity of \Cref{eq:mi_3inner}.

\section{Implementation details}
\label{apdx:imp_detail}

In this Section, we provide additional technical details of \gls{MINDE}. We discuss the different variants of our method their implementation details, including detailed information about the \gls{MI} estimators alternatives considered in \Cref{sec:experiments}.

\begin{algorithm}[h] 

\DontPrintSemicolon
\SetAlgoLined
\SetNoFillComment
\SetKwInOut{Parameter}{parameter}
\LinesNotNumbered 
\caption{\gls{MINDE}--\textsc{c} (Single Training Step) }
\label{algo:mind_c_training}
\KwData{$ [X_0 , Y_0] \sim \mu^C $ }  
\Parameter{$net_\theta( )$, with $\theta$ current parameters}
$t \sim \mathcal{U}[0,T]$ \tcp*{Importance sampling can be used to reduce variance} 
$ X_t  \gets k_tX_0+\left(k^2_t\int_0^t k^{-2}_sg^2_{s}\dd s\right)^{\frac{1}{2}} \epsilon$, with $\epsilon\sim \gamma_1$ \tcp*{ r.h.s. of \Cref{eq:diffsdesjoint}, diffuse the variable $X$ to timestep $t$} 

$c\sim \text{Bernoulli}(d)$ \tcp*{Sample binary variable $c$ with probability $d$}

\uIf{$c = 0 $ }{
$\frac{\hat{\epsilon}}{\left(k^2_t\int_0^t k^{-2}_sg^2_{s}\dd s\right)^{\frac{1}{2}}} \gets net_\theta([X_t , 0], t, c = 0 ) $ \tcp*{Estimated unconditional score }
}
\uElse{ 
$\frac{\hat{\epsilon}}{\left(k^2_t\int_0^t k^{-2}_sg^2_{s}\dd s\right)^{\frac{1}{2}}}  \gets net_\theta([X_t , Y_0], t, c = 1 ) ) $ \tcp*{ Estimated conditional score} 
}
$L=\frac{g^2_t}{\left(k^2_t\int_0^t k^{-2}_sg^2_{s}\dd s\right)}\norm{\epsilon-\hat{\epsilon}}^2$ \tcp*{ Compute Montecarlo sample associated to \Cref{eq:loss}} 
\Return Update $\theta$ according to gradient of $L$
\end{algorithm}

\begin{algorithm}[h] 

\DontPrintSemicolon
\SetAlgoLined
\SetNoFillComment
\SetKwInOut{Parameter}{parameter}
\LinesNotNumbered 
\caption{\gls{MINDE}--\textsc{c} }
\label{algo:mind_c}
\KwData{$ [X_0 , Y_0] \sim \mu^C $ }  
\Parameter{$\sigma$, $option$ }
$t \sim \mathcal{U}[0,T]$ \tcp*{Importance sampling can be used to reduce variance} 
$ X_t  \gets k_tX_0+\left(k^2_t\int_0^t k^{-2}_sg^2_{s}\dd s\right)^{\frac{1}{2}} \epsilon$, with $\epsilon\sim \gamma_1$ \tcp*{ r.h.s. of \Cref{eq:diffsdesjoint}, diffuse the variable $X$ to timestep $t$} 
$\tilde{s}_t^{\mu^A} \gets net_\theta([X_t , 0], t, c = 0 ) $ \tcp*{Use the unique score network to  compute }
$\tilde{s}_t^{\mu^A{_{Y_0}}} \gets net_\theta([X_t , Y_0], t, c = 1 ) ) $ \tcp*{ marginal and conditional scores}

\uIf{$option = 1 $ }{ 
  $\hat{I} \gets  T \frac{g^2_t }{2} \norm{{\tilde s}^{\mu^{A}}_{t} - \tilde{s}_t^{\mu^{A_{Y_0}}}  }^2  $
}
\uElse{ 
    $\chi_t \gets  \left(k^2_t\sigma^2+k^2_t\int_0^t k^{-2}_sg^2_{s}\dd s\right) $ 
    \\
    $\hat{I}  \gets T \frac{g^2_t}{2}   \left [\norm{{\tilde s}^{\mu^{A}}_{t} + \frac{X_t }{\chi_t}  }^2  
    -    \norm{\tilde{s}_t^{\mu^{A_{Y_0}}}  +\frac{X_t}{\chi_t} }^2  
     \right ]
    $
}
\Return $\hat{I}$
\end{algorithm}

\subsection{\gls{MINDE}-c}
\label{apdx:minde_c}
In all experiments, we consider the first variable as the main variable and the second variable as the conditioning signal. A single neural network is used to model the conditional and unconditional score. It accepts as inputs the two variables, the diffusion time $t$, and an additionally binary input $c$ which enable the conditional mode. To enable the conditional mode, we set $c=1$ and feed the network with both the main variable and the conditioning signal, obtaining  $\tilde{s}_t^{\mu^A{_{Y_0}}} $. To obtain the marginal score $\tilde{s}_t^{\mu^A} $, we set $c=0$ and the conditioning signal is set to zero value.

A randomized procedure is used for training. For each training step, with probability $d$, the main variable is diffused and the score network is fed with the diffused variable, the conditioning variable, the diffusion time signal and the conditioning signal is set to $c=1$.
On the contrary, with probability $1-d$, to enable the network to learn the unconditional score, the network is fed only with the diffused modality, the diffusion time and $c = 0$. In contrast to the first case, the conditioning is not provided to the score network and replaced with a zero value vector. Pseudocode is presented in \Cref{algo:mind_c_training}.

Actual estimation of the \gls{MI} is then possible either by leveraging \Cref{eq:mi_2} or \Cref{eq:mi_1}, referred to in the main text as difference \textit{outside} or \textit{inside} the score respectively (\gls{MINDE}\--\textsc{c}($\sigma$), \gls{MINDE}\--\textsc{c}). A pseudo-code description is provided in \Cref{algo:mind_c}.

\subsection{\gls{MINDE}-j}
\label{apdx:minde_j}
The joint variant of our method, \gls{MINDE}-\textsc{j} is based on the parametrized joint processes in \Cref{eq:maskeddiffusion}. Also in this case, instead of training a separate score network for each possible combination of conditional modalities, we use a single architecture that accepts both variables, the diffusion time $t$ and the coefficients $\alpha,\beta$. This approach allows modelling the joint score network ${\tilde s}^{\mu^{C}}_{t}$
by setting $\alpha = \beta = 1$. Similarly, to obtain the conditional scores it is sufficient to set $\alpha=1 ,\beta = 0$ or $\alpha=0 ,\beta = 1$, corresponding to $\tilde{s}_t^{\mu^A_{Y_0}}$ and $\tilde{s}_t^{\mu^A_{X_0}}$ respectively. 

Training is carried out again through a randomized procedure. At each training step, with probability $d$, both variables are diffused. In this case, the score network is fed with diffusion time $t$, along with $X_t,Y_t$ and the two parameters $\alpha = \beta = 1$. 
With probability $1-d$, instead, we randomly select one variable to be diffused, while we keeping constant the other. For instance, if $A$ is the one which is diffused, we set $\alpha =1$ and $\beta = 0$. Further details are presented in \Cref{algo:mind_j_training}.

Once the score network is trained, \gls{MI} estimation can be obtained following the procedure explained in \Cref{algo:mind_j}. Two options are possible, either by computing the difference between the parametric scores 
outside the same norm (\Cref{eq:mi_3} \gls{MINDE}\--\textsc{j}($\sigma$) or inside (\Cref{eq:mi_3inner} \gls{MINDE}\--\textsc{j}). Similarly to the conditional case, an \textit{option} parameter can be used to switch among the two. 

\begin{algorithm}[h] 

\DontPrintSemicolon
\SetAlgoLined
\SetNoFillComment
\SetKwInOut{Parameter}{parameter}
\LinesNotNumbered 
\caption{\gls{MINDE}--\textsc{j}  (Single Training Step) }
\label{algo:mind_j_training}
\KwData{$ [X_0 , Y_0] \sim \mu^C $ }  
\Parameter{$net_\theta( )$, with $\theta$ current parameters}
$t \sim \mathcal{U}[0,T]$ \tcp*{Importance sampling can be used to reduce variance} 

$ [X_t , Y_t] \gets k_t[X_0,Y_0]+\left(k^2_t\int_0^t k^{-2}_sg^2_{s}\dd s\right)^{\frac{1}{2}} [\epsilon_1,\epsilon_2]$, with $\epsilon_{1,2}\sim \gamma_1$  \tcp*{l.h.s. \Cref{eq:diffsdesjoint}, diffuse modalities to timestep $t$}

$c\sim \text{Bernoulli}(d)$ \tcp*{Sample binary variable $c$ with probability $d$}

\uIf{$c = 0 $ }{
$\frac{ [\hat\epsilon_1,\hat\epsilon_2]}{\left(k^2_t\int_0^t k^{-2}_sg^2_{s}\dd s\right)^{\frac{1}{2}}} \gets net_\theta([X_t , Y_t], t, [1,1])$ \tcp*{Estimated unconditional score }

$L=\frac{g^2_t}{\left(k^2_t\int_0^t k^{-2}_sg^2_{s}\dd s\right)}\norm{[\epsilon_1,\epsilon_2]-[\hat{\epsilon_1},\hat{\epsilon_2}]}^2$ \tcp*{ Compute Montecarlo sample associated to \Cref{eq:loss}}

}
\uElse{ 
\uIf{\text{Bernoulli}(0.5)}
{$\frac{ \hat\epsilon_1}{\left(k^2_t\int_0^t k^{-2}_sg^2_{s}\dd s\right)^{\frac{1}{2}}} \gets net_\theta([X_t , Y_0], t, [1,0])$ \tcp*{Estimated Conditional score }}
$L=\frac{g^2_t}{\left(k^2_t\int_0^t k^{-2}_sg^2_{s}\dd s\right)}\norm{\epsilon_1-\hat{\epsilon_1}}^2$\\
\uElse{
$\frac{ \hat\epsilon_2}{\left(k^2_t\int_0^t k^{-2}_sg^2_{s}\dd s\right)^{\frac{1}{2}}} \gets net_\theta([X_0 , Y_t], t, [0,1])$ \tcp*{Estimated Conditional score }
$L=\frac{g^2_t}{\left(k^2_t\int_0^t k^{-2}_sg^2_{s}\dd s\right)}\norm{\epsilon_2-\hat{\epsilon_2}}^2$
 }
}

\Return Update $\theta$ according to gradient of $L$

\end{algorithm}

\begin{algorithm}[h] 

\DontPrintSemicolon
\SetAlgoLined
\SetNoFillComment
\SetKwInOut{Parameter}{parameter}
\LinesNotNumbered 
\caption{\gls{MINDE}--\textsc{j} }
\label{algo:mind_j}
\KwData{$ [X_0 , Y_0] \sim \mu^C $ }  
\Parameter{$\sigma$, $option$ }
$t \sim \mathcal{U}[0,T]$ \tcp*{Importance sampling can be used to reduce variance} 

$ [X_t , Y_t] \gets k_t[X_0,Y_0]+\left(k^2_t\int_0^t k^{-2}_sg^2_{s}\dd s\right)^{\frac{1}{2}} [\epsilon_1,\epsilon_2]$, with $\epsilon_{1,2} \sim \gamma_1$  \tcp*{l.h.s. \Cref{eq:diffsdesjoint}, diffuse modalities to timestep $t$} 
$\tilde{s}_t^{\mu^C} \gets net_\theta([X_t , Y_t], t, [1,1]) $ \tcp*{Use the unique score network to  compute joint}
$\tilde{s}_t^{\mu^A_{Y_0}} \gets net_\theta([X_t , Y_0], t, [1,0]) $ \tcp*{ and conditional scores} 
$\tilde{s}_t^{\mu^A_{X_0}} \gets net_\theta([X_0 , Y_t], t, [0,1]) $\\

\uIf{$option = 1 $ }{ 
  $\hat{I} \gets  T \frac{g^2_t }{2} \norm{{\tilde s}^{\mu^{C}}_{t}-[\tilde s^{\mu^{A_{Y_0}}}_{t},\tilde s^{\mu^{B_{X_0}}}_{t}] }^2  $
}
\uElse{ 
    $\chi_t \gets  \left(k^2_t\sigma^2+k^2_t\int_0^t k^{-2}_sg^2_{s}\dd s\right) $ 
    \\
    $\hat{I}  \gets T \frac{g^2_t}{2}   \left [\norm{{\tilde s}^{\mu^{C}}_{t} + \frac{[X_t , Y_t]}{\chi_t}  }^2  
    -    \norm{\tilde s^{\mu^{A_{Y_0}}}_{t}  + \frac{X_t}{\chi_t} }^2  
    -    \norm{\tilde s^{\mu^{B_{X_0}}}_{t} +  \frac{Y_t}{\chi_t}  }^2 \right ]
    $
}
\Return $\hat{I}$
\end{algorithm}

\subsection{Technical settings for \gls{MINDE}-c and \gls{MINDE}-j }

We follow the implementation of \cite{bounoua2023multimodal} which uses stacked multi-layer perception (MLP) with skip connections. We adopt a simplified version of the same score network architecture: this involves three Residual MLP blocks. We use the \textit{Adam optimizer} \citep{kingma2014adam} for training and Exponential moving average (EMA) with a momentum parameter $m = 0.999$. We use importance sampling at train and test-time. We returned the mean estimate on the test data set over 10 runs. 

The hyper-parameters are presented in \Cref{table:minde_j} and \Cref{table:minde_c} for \gls{MINDE}-\textsc{j} and \gls{MINDE}-\textsc{c} respectively. Concerning the consistency tests (\Cref{sec:consistency}), we independently train an autoencoder for each version of the \textsc{mnist} dataset with $r$ rows available. 

\renewcommand{\tabcolsep}{2.0pt}

\begin{table}[h]
\caption{\gls{MINDE}-\textsc{j} score network training hyper-parameters. $Dim$ of the task correspond the sum of the two variables dimensions, whereas $d$ corresponds to the randomization probability.}
\centering
\begin{tabular}{ccccccccc}
\toprule
 & $ d $ & Width &Time embed & Batch size & Lr & Iterations & Number of params  \\
\midrule 
Benchmark ($Dim \leq 10$) & 0.5  & 64 &64& 128 &  1e-3 & 234k& 55490\\
Benchmark ($Dim = 50$)  & 0.5   & 128 &128& 256 &  2e-3 & 195k& 222100\\
Benchmark ($Dim =100$) & 0.5  & 256 &256& 256 &  2e-3 & 195k & 911204\\ 
\midrule
Consistency tests & 0.5 & 256 & 256 & 64 &   1e-3 & 390k & 1602080\\
\bottomrule
\end{tabular}
\label{table:minde_j}
\end{table}

\renewcommand{\tabcolsep}{2.0pt}

\begin{table}[h]
\caption{\gls{MINDE}-\textsc{c} score network training hyper-parameters. $Dim$ of the task correspond the sum of the two variables dimensions, and $d$ corresponds to the randomization probability.}
\centering
\begin{tabular}{ccccccccc}
\toprule
 & $ d $ &  Width &Time embed & Batch size & Lr & Iterations & Number of params  \\
\midrule 
Benchmark ($Dim \leq 10$) & 0.5   & 64 & 64 & 128 &  1e-3 & 390k&55425  \\
Benchmark ($Dim = 50$)  & 0.5   & 128 & 128 & 256 &  2e-3 & 290k&220810 \\
Benchmark ($Dim =100$) & 0.5   & 256 & 256 & 256 &  2e-3 & 290k & 898354 \\ 
\midrule
Consistency tests & 0.5  & 256 & 256 & 64 &   1e-3 &390k& 1597968 \\
\bottomrule
\end{tabular}
\label{table:minde_c}
\end{table}

\subsection{Neural estimators implementation}
We use the package \textit{benchmark-mi}\footnote{\url{https://github.com/cbg-ethz/bmi}} implementation to study the neural estimators.
We use MLP architecture with 3 layers of the same width as in \gls{MINDE}. We use the same training procedure as in \cite{czyz2023beyond}, including early stopping strategy. We return the highest estimate on the test data.

\section{Ablations study}
\label{apdx:sigma_ablations}

\subsection{ $\sigma$ Ablation study }
We hereafter report in \Cref{table_complete} the results of all the variants of \gls{MINDE}, including different values of $\sigma$ parameter. For completeness in our experimental campaign, we report also the results of non neural competitors, similarly to the work in \cite{czyz2023beyond}. In summary, the \gls{MINDE}\--\textsc{c}/\textsc{j} versions (``\textit{difference inside}'') of our estimator prove to be more robust than the \gls{MINDE}\--\textsc{c}/\textsc{j}($\sigma$) (``\textit{difference outside}'') counterpart, especially for the joint variants. Nevertheless, it is interesting to notice that the ``\textit{difference outside}'' variants are stable and competitive against a very wide range of values of $\sigma$ (ranging from $0.5$ to $10$), with their best value typically achieved for $\sigma=1.0$.

\begin{landscape}

\begin{table}
\tiny

\begin{tabular}{lrrrrrrrrrrrrrrrrrrrrrrrrrrrrrrrrrrrrrrrr}

\textbf{GT} & 0.2 & 0.4 & 0.3 & 0.4 & 0.4 & 0.4 & 0.4 & 1.0 & 1.0 & 1.0 & 1.0 & 0.3 & 1.0 & 1.3 & 1.0 & 0.4 & 1.0 & 0.6 & 1.6 & 0.4 & 1.0 & 1.0 & 1.0 & 1.0 & 1.0 & 1.0 & 1.0 & 1.0 & 1.0 & 0.2 & 0.4 & 0.2 & 0.3 & 0.2 & 0.4 & 0.3 & 0.4 & 1.7 & 0.3 & 0.4 \\
\bottomrule
\end{tabular}
\caption{ \gls{MINDE}--j and \gls{MINDE}--c $\sigma$ ablations study .Mean MI estimates over 10 seeds using N = 10000 samples compared each against the ground-truth. Color indicates relative
negative bias (red) and positive bias (blue). Our method and all neural estimators were trained with 100k training samples. List of abbreviations ( \textit{Mn}: Multinormal,  \textit{St}: Student-t, \textit{Nm}: Normal, \textit{Hc}: Half-cube, \textit{Sp}: Spiral)
} \label{table_complete}
\end{table}
\end{landscape}

\subsection{ Full results with standard deviation }
\begin{figure}[H]
    \centering
    \begin{subfigure}{0.49\textwidth}
    \includegraphics[width=1\textwidth]{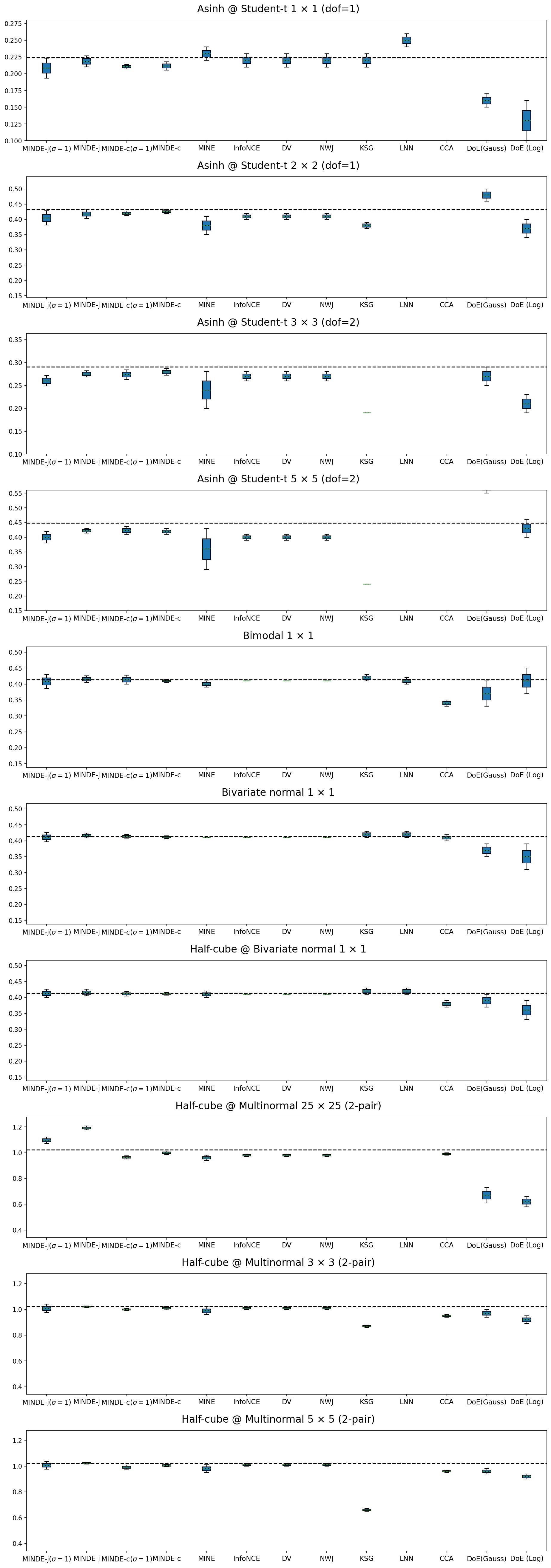}
    \end{subfigure}
    \begin{subfigure}{0.49\textwidth}
    \centering
      \includegraphics[width=1\textwidth]{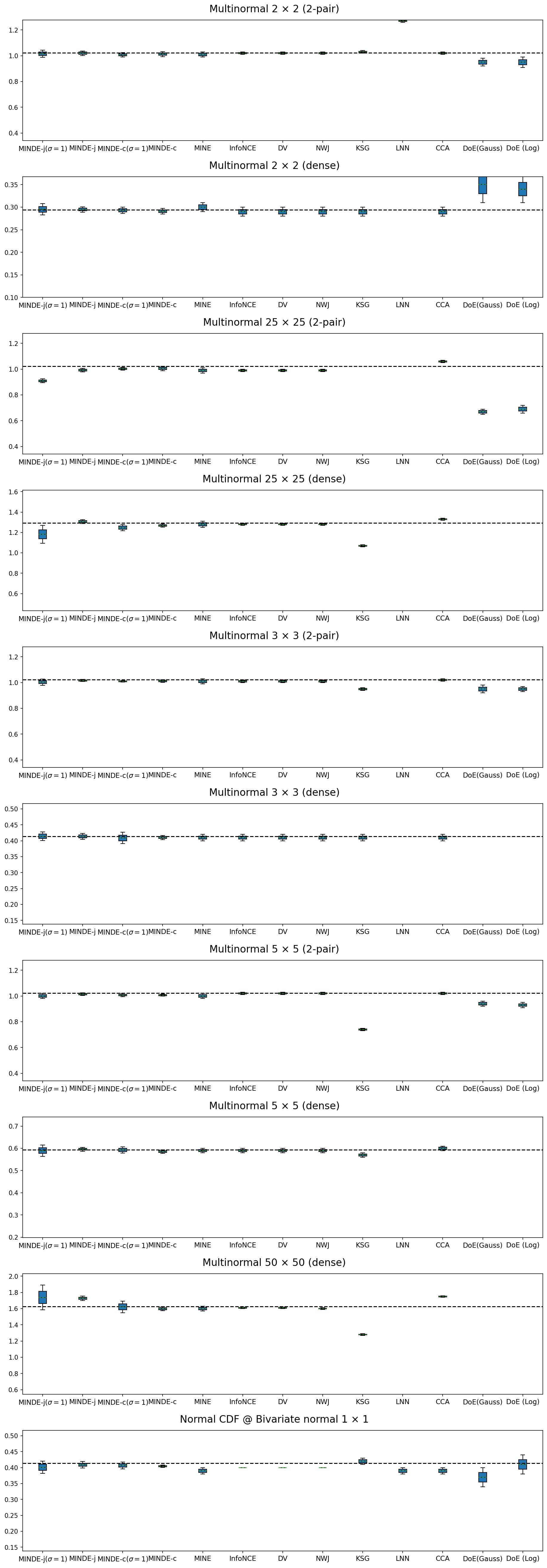}
    \end{subfigure}
    \caption{We report MI estimate results over 10 seeds for N =10000 for our method and competitors for training size 100k sample. A method absent from the depiction implies either non convergence during training or results out of scale}
    \label{box1}
\end{figure}

\begin{figure}[H]
    \centering
    \begin{subfigure}{0.49\textwidth}
    \includegraphics[width=1\textwidth]{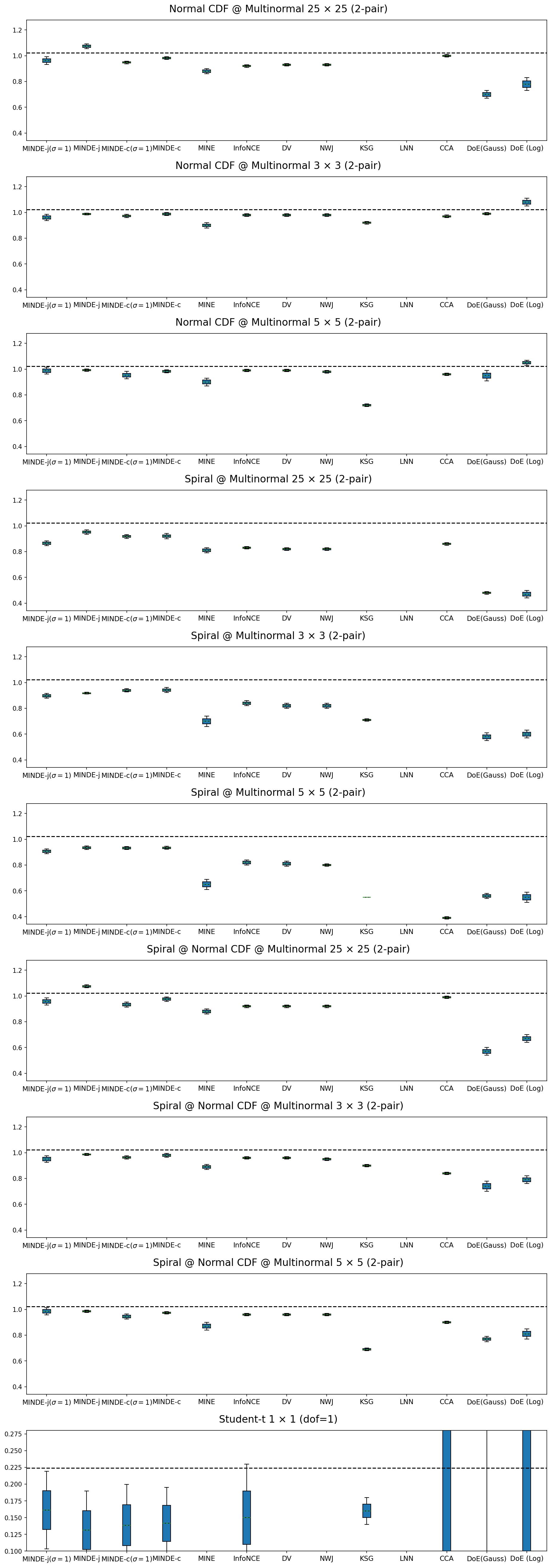}
    \end{subfigure}
    \begin{subfigure}{0.49\textwidth}
    \centering
      \includegraphics[width=1\textwidth]{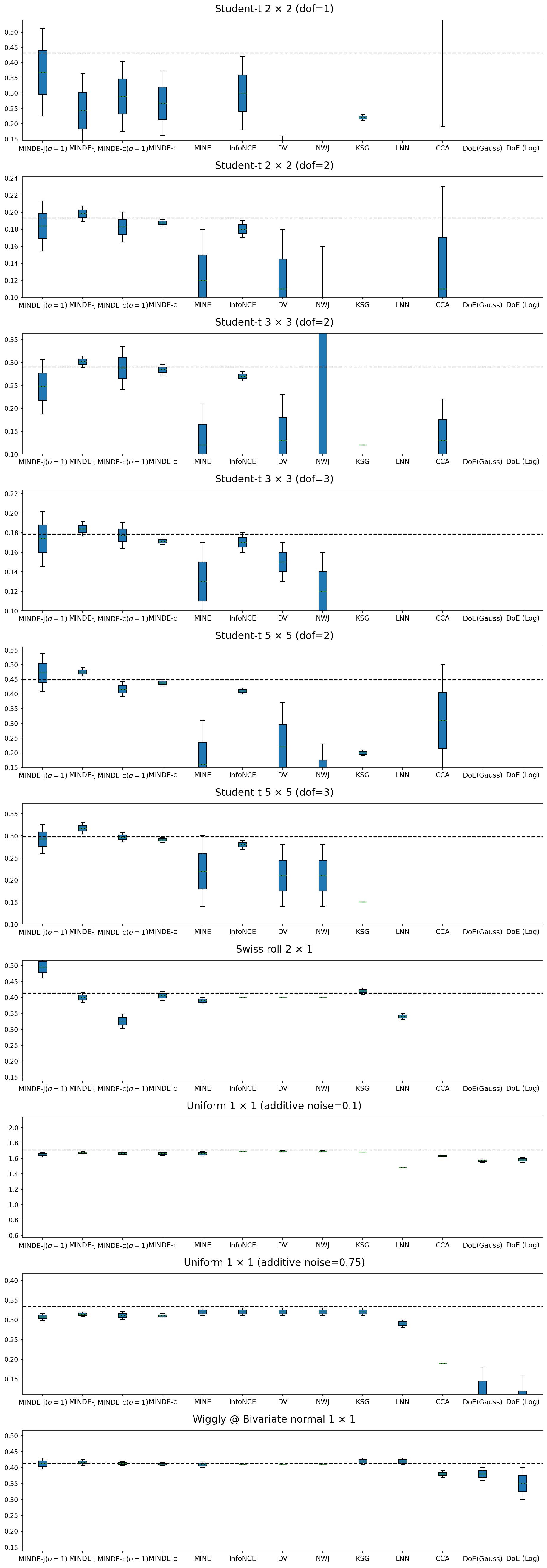}
    \end{subfigure}
     \caption{We report MI estimate results over 10 seeds for N =10000 for our method and competitors for training size 100k sample.}
     \label{box2}
\end{figure}

We report in \Cref{mean_100k} mean results without quantization for the different methods. \Cref{box1,box2} contains box-plots for all the competitors and all the tasks. 

\renewcommand{\tabcolsep}{0.0pt}

\begin{landscape}

\begin{table}
\tiny
\begin{tabular}{lrrrrrrrrrrrrrrrrrrrrrrrrrrrrrrrrrrrrrrrr}
\toprule

\textbf{GT} & 0.22 & 0.43 & 0.29 & 0.45 & 0.41 & 0.41 & 0.41 & 1.02 & 1.02 & 1.02 & 1.02 & 0.29 & 1.02 & 1.29 & 1.02 & 0.41 & 1.02 & 0.59 & 1.62 & 0.41 & 1.02 & 1.02 & 1.02 & 1.02 & 1.02 & 1.02 & 1.02 & 1.02 & 1.02 & 0.22 & 0.43 & 0.19 & 0.29 & 0.18 & 0.45 & 0.30 & 0.41 & 1.71 & 0.33 & 0.41 \\
\bottomrule
\end{tabular}

\caption{Mean estimate over 10 seeds using N = 10000 samples compared each against the ground-truth. \\Our method and all neural estimators were trained with 100k training samples. \\List of abbreviations ( \textit{Mn}: Multinormal,  \textit{St}: Student-t, \textit{Nm}: Normal, \textit{Hc}: Half-cube, \textit{Sp}: Spiral)
}
\label{mean_100k}
\end{table}

\end{landscape}

\subsection{Training size ablation study }
\begin{figure} [H]
   \centering
   \includegraphics[width=1\textwidth]{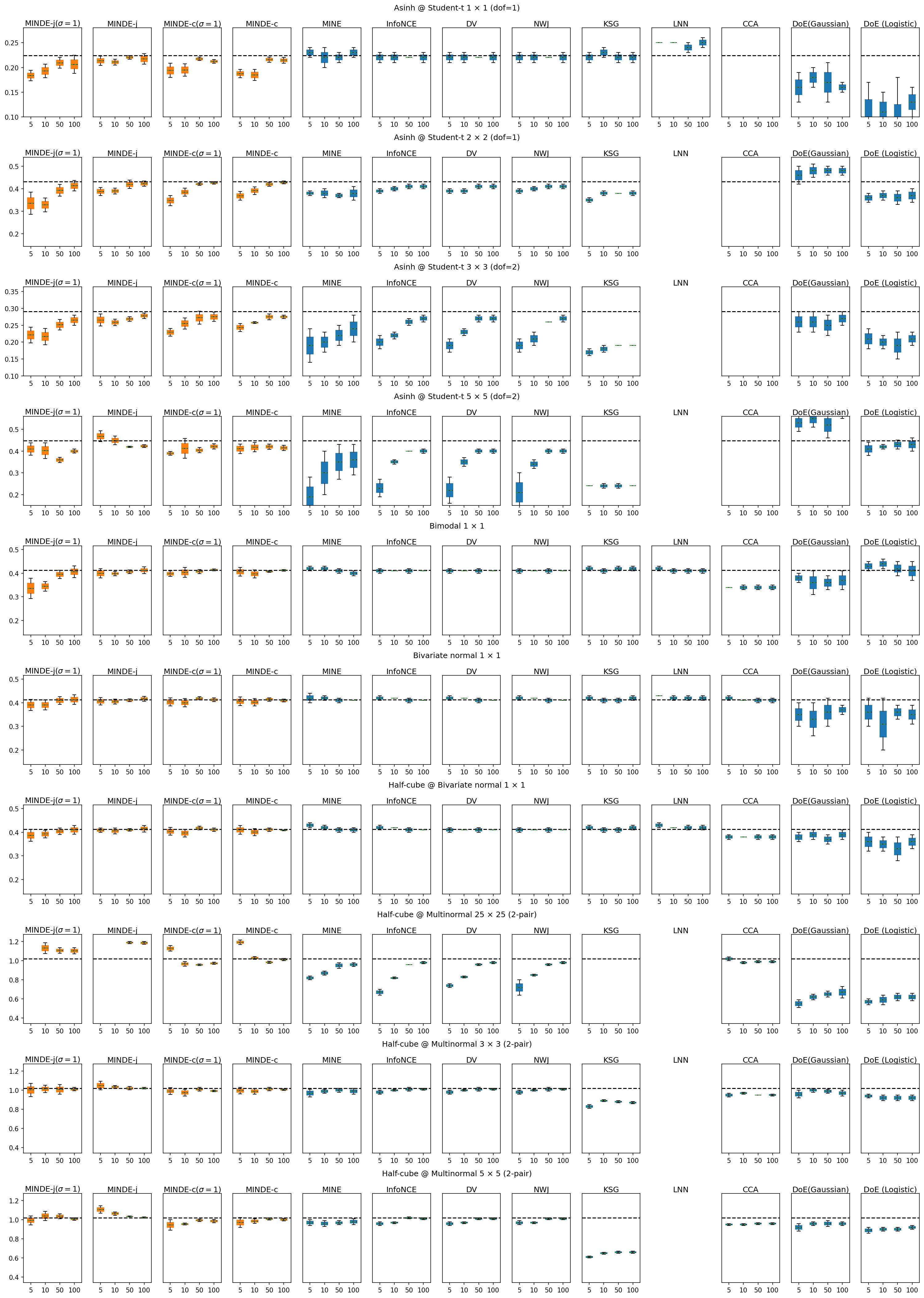}
    \caption{Training Size ablation study : We report MI estimate results for our method and competitors as a function of the training size used (5k,10k,50k,100k).
    For readability, we discard the baselines with estimation (error $>$ 2 * GT) or high standard deviation. All results are averaged   over 5 seeds. Due the benchmark size, we split the results into 4 figures each containing 10 benchmarks. A method absent from the depiction implies either non convergence during training or results out of scale. In this first plot we report tasks 1-10. }
       \label{boxplots}
\end{figure}

\begin{figure}[H]
   \centering
   \includegraphics[width=1\textwidth]{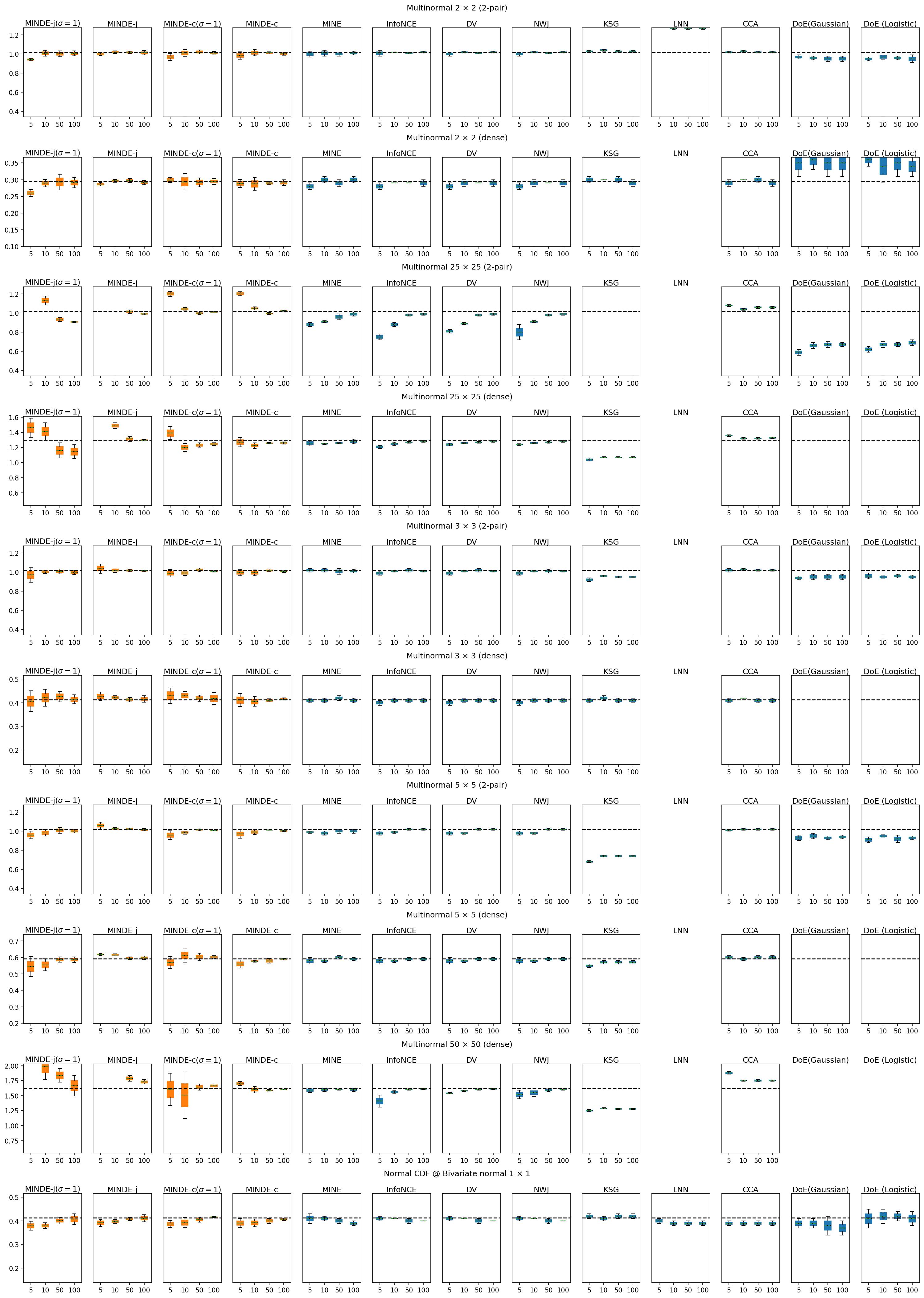}       
    \caption{{Part 2 of \Cref{boxplots}, tasks 11-20. }}
    \label{boxplots2}
\end{figure}

\begin{figure}[H]
   \centering
   \includegraphics[width=1\textwidth]{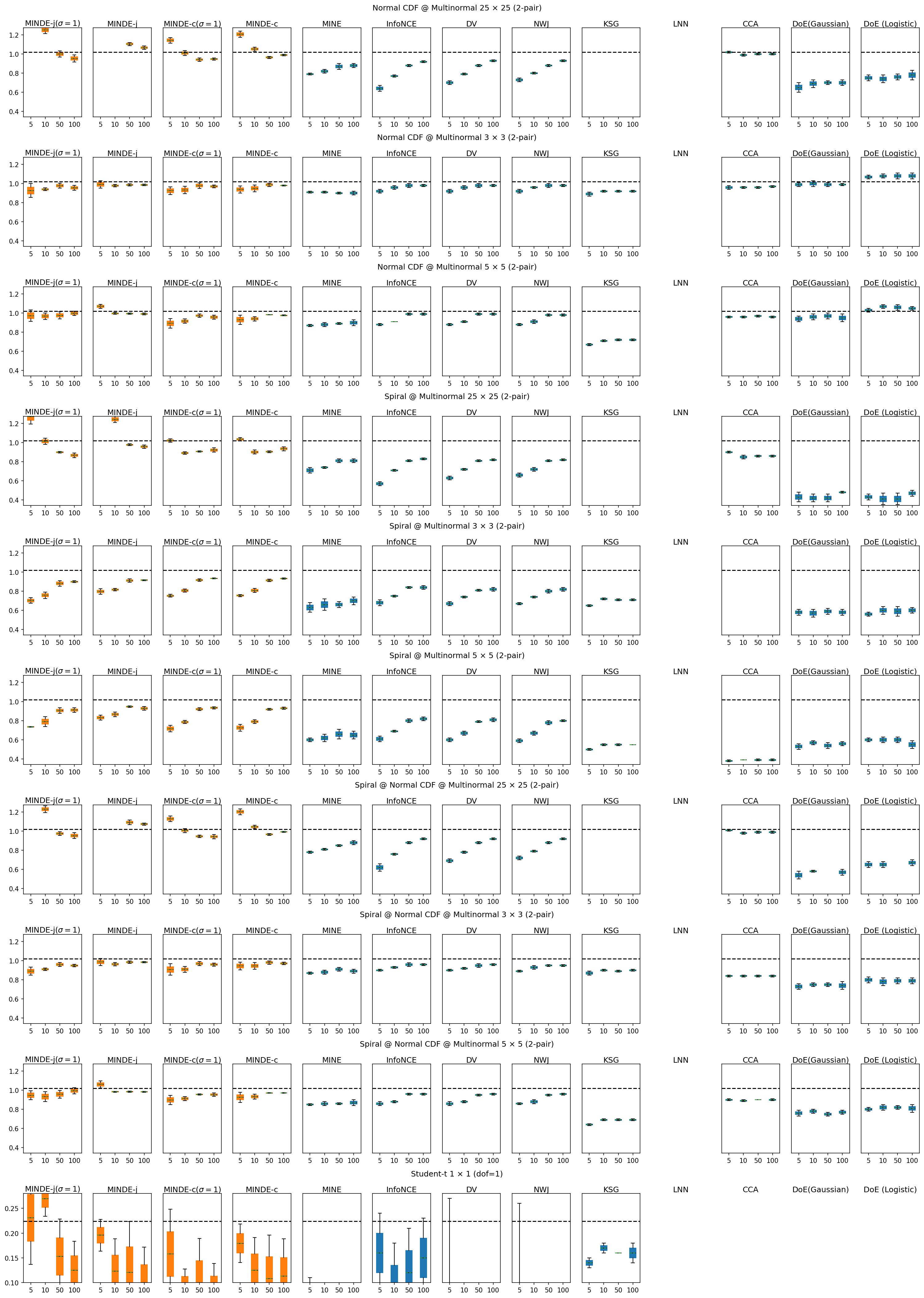}       
     \caption{{Part 3 of \Cref{boxplots}, tasks 21-30.
     \label{boxplots3}}}
\end{figure}

\begin{figure}[H]
   \centering
   \includegraphics[width=1\textwidth]{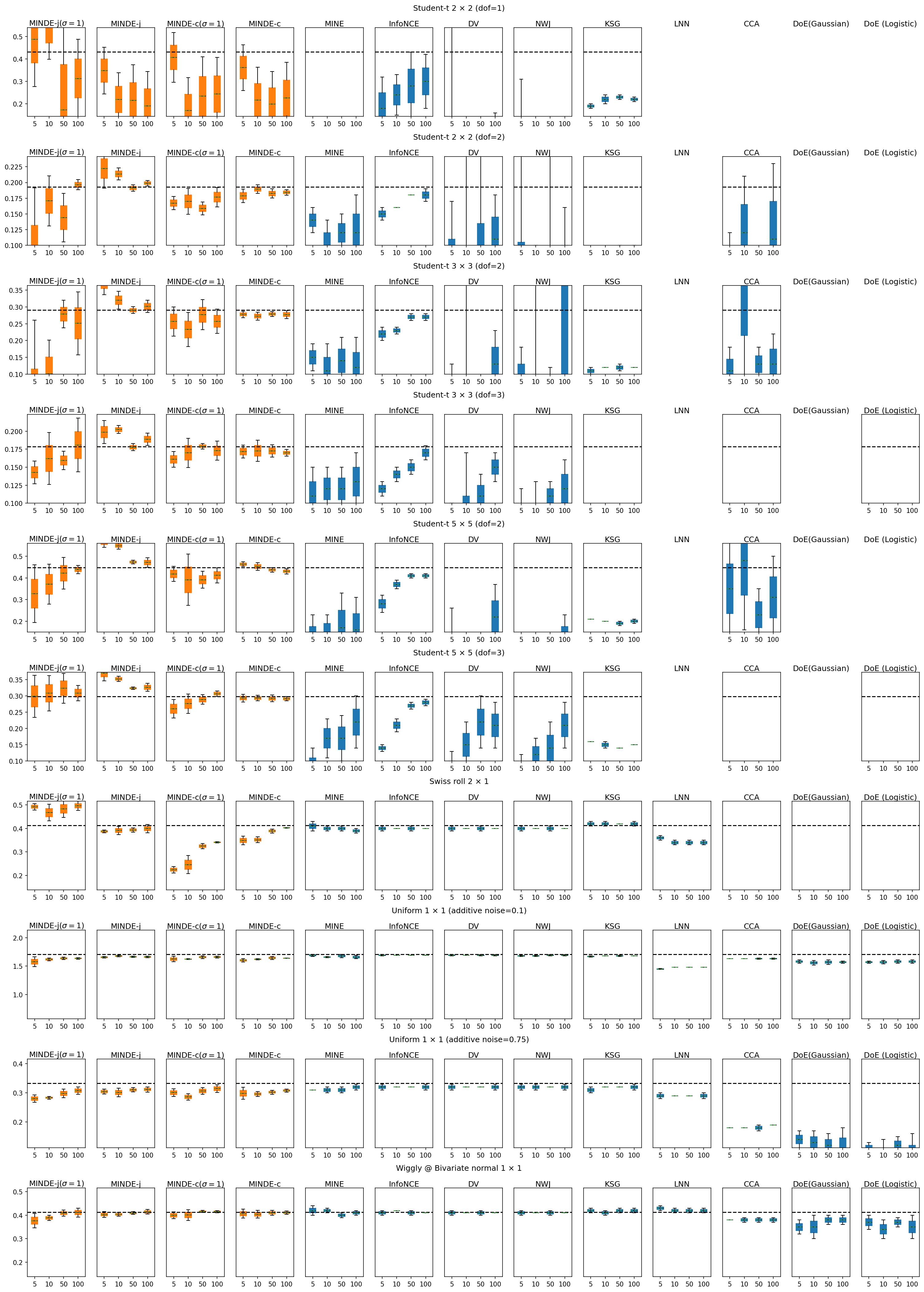}       
    \caption{{Part 4 of \Cref{boxplots}, tasks 31-40.
    \label{boxplot4}}}
 
\end{figure}

We here report, in \Cref{boxplots,boxplots2,boxplots3,boxplot4} the results of our ablation study on the training size, varying in the range 5k,10k,50k,100k.

\section{Analysis of conditional diffusion dynamics using \gls{MINDE}}
Diffusion models have achieved outstanding success in generating high-quality images, text, audio, and video across various domains. Recently, the generation of diverse and realistic data modalities (images, videos, sound) from open-ended text prompts \citep{ramesh2022, saharia2022, rombach2022} has projected practitioners into a whole new paradigm for content creation. A remarkable property of our \gls{MINDE} method is its generalization to any score based model. Then, our method can be considered as a plug and play tool to explore information theoretic properties of score-based diffusion models: in particular, in this section we use \gls{MINDE} to estimate \gls{MI} in order to explain the dynamics of image conditional generation, by analyzing the influence of the prompt on the image generation through time.


\paragraph{Prompt influence of conditional sampling.}

Generative diffusion models can be interpreted as \textit{iterative} schemes in which starting from pure noise, at each iteration, refinements are applied until a sample from the data distribution is obtained. 
In recent work on \textbf{text conditional image generation} (image $A$, text prompt $B$) by \citet{balaji2022ediffi}, it has been observed that the role of the text prompt throughout the generative process has not constant importance . Indeed: \textit{``At the early sampling stage, when the input data to the denoising network is closer to the random noise, the diffusion model mainly relies on the text prompt to guide the sampling process. As the generation continues, the model gradually shifts towards visual features to denoise images, mostly ignoring the input text prompt''} \citep{balaji2022ediffi}. Such claim has been motivated by carefully engineered metric analysis such as self and cross attention maps between images and text, as a function of the generation time, as well as visual inspection of the change in generated images when switching the prompt at different stages of the refinement. 

Using \gls{MINDE}, we can refine heuristic-based methods and produce a similar analysis using theoretically sound information theoretic quantities. 
In particular, we analyze the conditional mutual information $\textsc{I}(A,B\g X_\tau)$, being $X_\tau$ the result of the generation process at time $\tau$ (recall that the time runs backward from $T$ to $0$ during generation, and consequently $A=X_0$ and $B=Y_0$). Such metric quantifies, given an observation of the generation process at time $\tau$, how much information the prompt $B$ carries about the final generated image $A$. Clearly, when $\tau=T$, the initial sample is independent from both $A$ and $B$. Consequently, the conditional mutual information will coincide with $\textsc{I}(A,B)$. 


More formally, we consider the following quantity:

\begin{flalign}\label{eq:sd_mi}
    \textsc{I}(A,B\g X_\tau) = \textsc{I}(A,B) - \left[  \textsc{I}(X_\tau,B) - \textsc{I}(X_\tau,A\g B)   \right ], \\
    = \textsc{I}(A,B) - \left[  H(X_\tau) -H(X_\tau|B) - H(X_\tau|B) + H(X_\tau |A,B)  \right ]\label{eq:sd_mi},\\
    = \textsc{I}(A,B) - \textsc{I}(X_\tau,B),
\end{flalign}

where \Cref{eq:sd_mi} is simplified due to the Markov chain $A- X_0-X_\tau$, so $H(X_\tau |A,B) = H(X_\tau |X_0,B) = H(X_\tau|B)$. Next, we use our \gls{MINDE} estimator, whereby the marginal and conditional entropies can be estimated efficiently. The following approximation of the quantity in interest can be derived:

\begin{align} 
 \label{eq:sd_mi_eq}
\textsc{I}(A,B\g X_\tau)  
\simeq \E_{\mathbb{P}^{\mu^{C}}}\left[\int\limits_0^\tau\frac{g^2_{t}}{2} \norm{
{\tilde s}^{\mu^{A}}_{t} (X_t) - \tilde{s}_t^{\mu^{A_{Y_0}}}  (X_t)
}^2 \dd t \right]
\end{align}

In our experiments, we also include a \gls{MINDE}-($\sigma$) version which  can be obtained similarly to \Cref{eq:sd_mi_eq}.

\paragraph{Experimental setting.}

We perform our experimental analysis of the influence of a prompt on image generation using Stable Diffusion \citep{rombach2022}, using the original code-base and pre-trained checkpoints.\footnote{\url{https://huggingface.co/stabilityai/stable-diffusion-2-1}} The original Stable Diffusion model was trained using the DDPM framework \citep{ho2020} on images latent space. This framework is equivalent to the discrete-time version of VPSDE \citep{song2021a}. Using the text prompt samples from \textsc{Laion} dataset \cite{schuhmann2022laion5b}, we synthetically generate image samples. We set guidance mechanism to 1.0 to ensure that the images only contain text conditional content. We use 1000 samples and approximate the integral using a Simpson integrator \footnote{\url{https://docs.scipy.org/doc/scipy/reference/generated/scipy.integrate.simpson.html}} with a discretization over 1000 timesteps.

\begin{figure}[H]
    \centering
    \begin{subfigure}{0.5 \textwidth}
        \includegraphics[page=1,width=\linewidth]{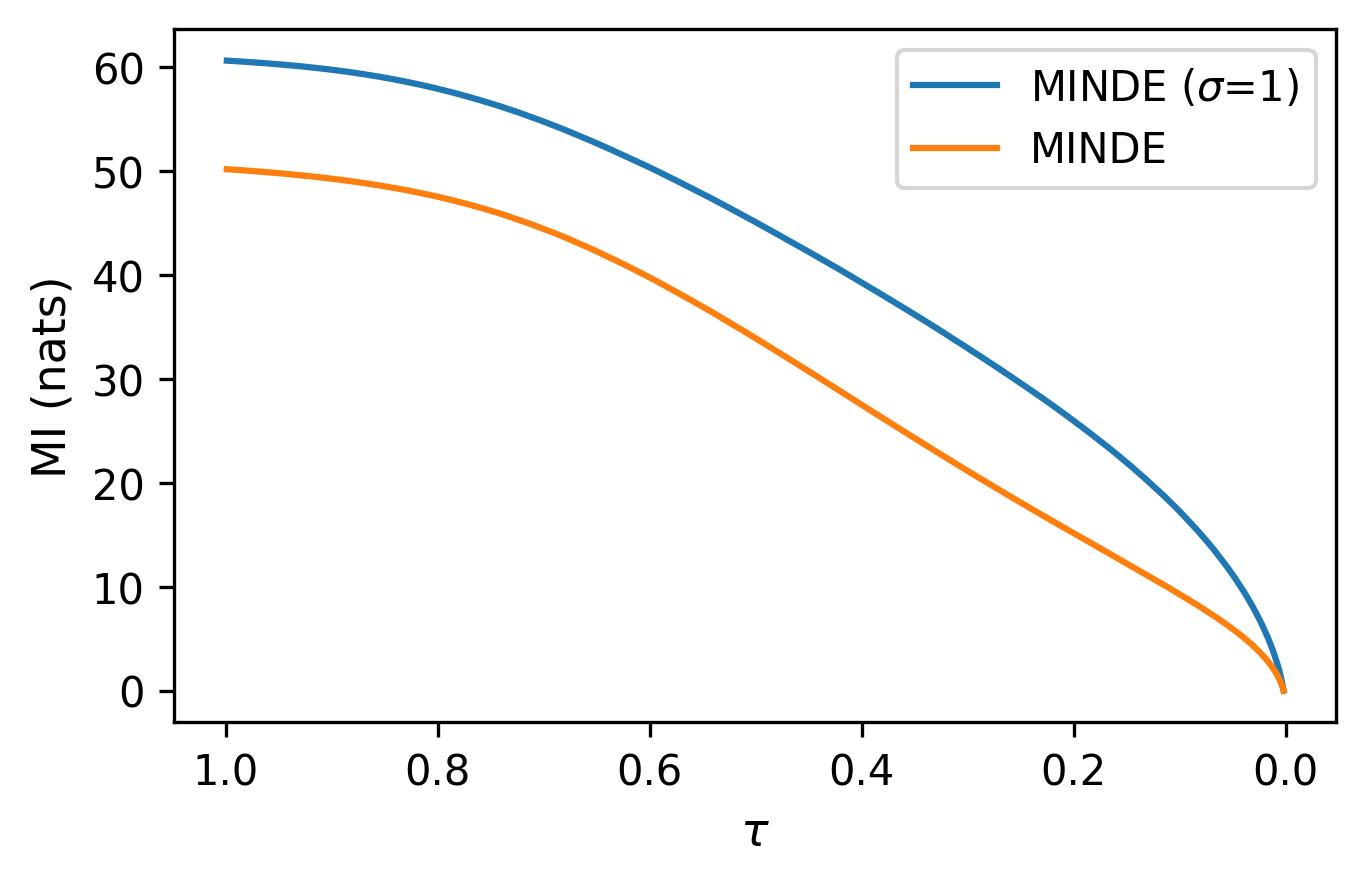}
        \label{fig:sd} 
    \end{subfigure}
    \caption{$I(A,B \g X_\tau)$  as a function of $\tau$. }
    \label{fig:sd_exp}
\end{figure}

\paragraph{Results.} 
We report in \Cref{fig:sd_exp} values of $\textsc{I}(A,B\g X_\tau)$ as a function of (reverse) diffusion time, where $A$ is in the image domain and $B$ is in the text domain. In a similar vein to what observed by \citet{balaji2022ediffi}, our results indicate that $\textsc{I}(A,B\g X_\tau)$ is very high when $\tau \simeq T$, which indicates that the text prompt has maximal influence during the early stage of image generation. This measurement is relatively stable at high \gls{MI} values until  $\tau \approx 0.8$. Then, the influence of the prompt gradually fades, as indicated by decreasing steadily \gls{MI} values. This corroborates the idea that mutual information can be adopted as an exploratory tool for the analysis of complex, high dimensional distributions in real use cases.

\begin{figure}[H]

    \centering
\begin{subfigure}{1.0 \textwidth}
    \includegraphics[page=1,width=\linewidth]
    {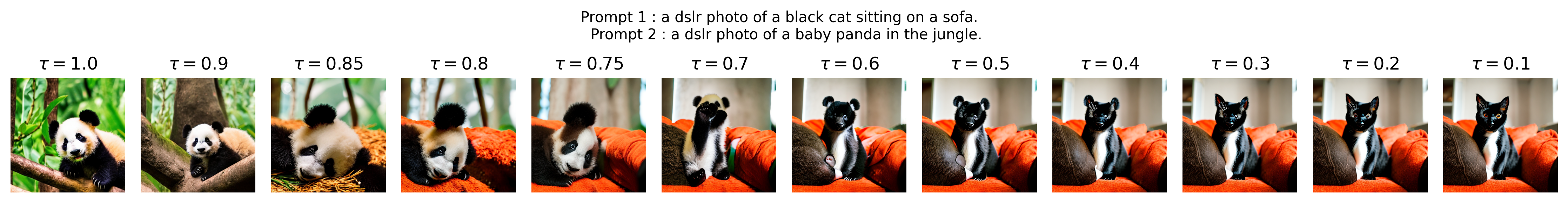}
\end{subfigure}

\begin{subfigure}{1.0 \textwidth}
    \includegraphics[page=1,width=\linewidth]{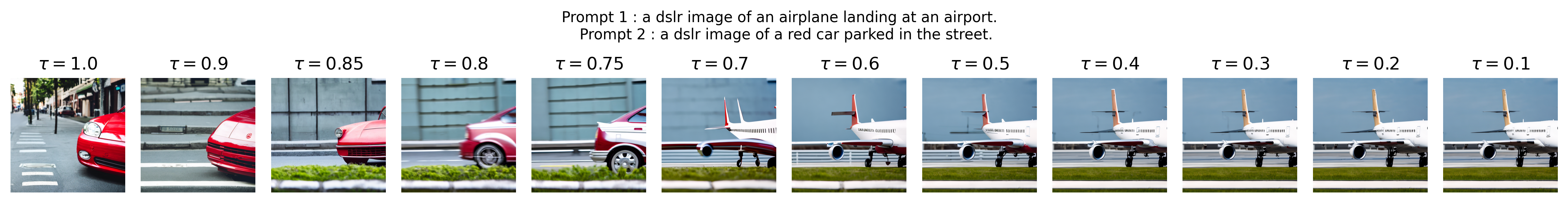}
\end{subfigure}

\begin{subfigure}{1.0 \textwidth}
    \includegraphics[page=1,width=\linewidth]{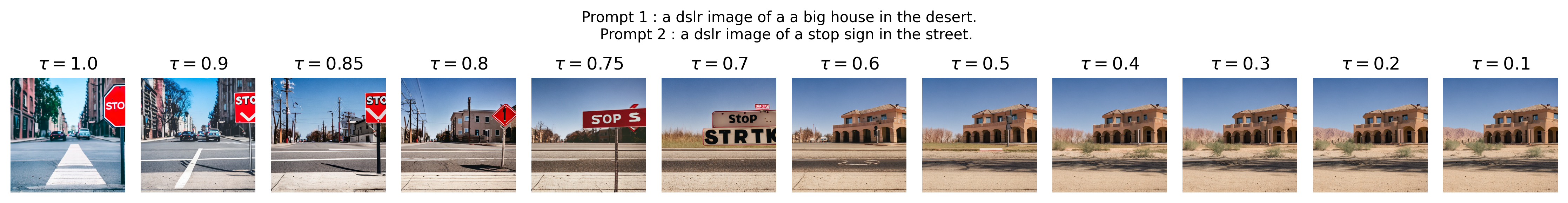}
\end{subfigure}

\begin{subfigure}{1.0 \textwidth}
    \includegraphics[page=1,width=\linewidth]{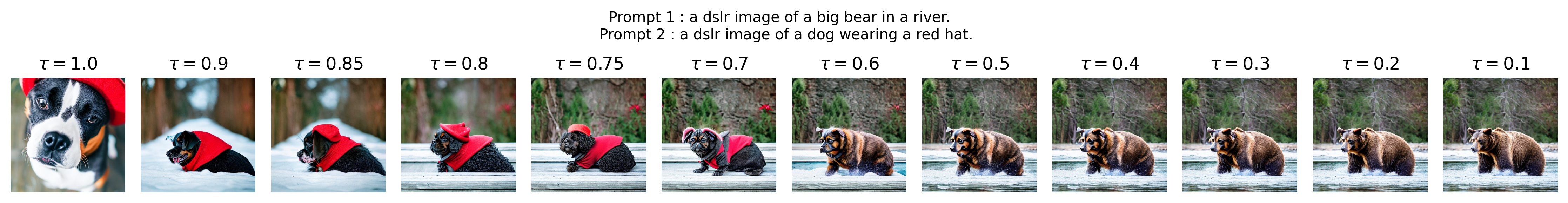}
\end{subfigure}

\caption{To validate the explanatory results obtained via the application of our \gls{MINDE} estimator, we perform the following experiment: Conditional generation is carried out with \textit{Prompt 1 } until time $\tau$, whereas after the conditioning signal is switched to \textit{Prompt 2}. We use the same Stable diffusion model as in the previous experiment with guidance scale set to 9. }
    \label{fig:qualiative_sd}
\end{figure}

The intuition pointed out by our \gls{MINDE} estimator is further consolidated by the qualitative samples in \Cref{fig:qualiative_sd}, where we perform the following experiment: we test whether switching from an original prompt to a different prompt during the backward diffusion semantically impacts the final generated images. We observe that changing the prompt before $\tau \simeq 0.8$ results almost surely with semantically coherent generated image with the second prompt.
Instead, when $\tau < 0.8$, the second prompt influence diminishes gradually. We observe that for all the qualitative samples shown in \Cref{fig:qualiative_sd} the second prompt has no influence on the generated image after $\tau < 0.7$.

\section{Scalabality of \gls{MINDE}}
In this Section, we study the generalization of our \gls{MINDE} estimator to more than two random variables. We consider the \textit{information interaction} between three random variables $A$,$B$ and $C$, defined as:

\begin{align}\label{3vars}
    \textsc{I}(A,B,C) = \textsc{I}(A,B) - \textsc{I} (A,B|C) \\
    = H(A) - H(A|B) - ( H(A|B,C) - H(A,C))
\end{align}
Estimation of such quantity is possible through a simple extension of \Cref{eq:maskeddiffusion} to three random variables, considering three parameters $\alpha,\beta,\gamma \in \{ 0,1\}$.

\begin{figure}[H]
\centering
\begin{subfigure}{0.40\textwidth}
    \centering
        \includegraphics[page=0.25,width=\linewidth]{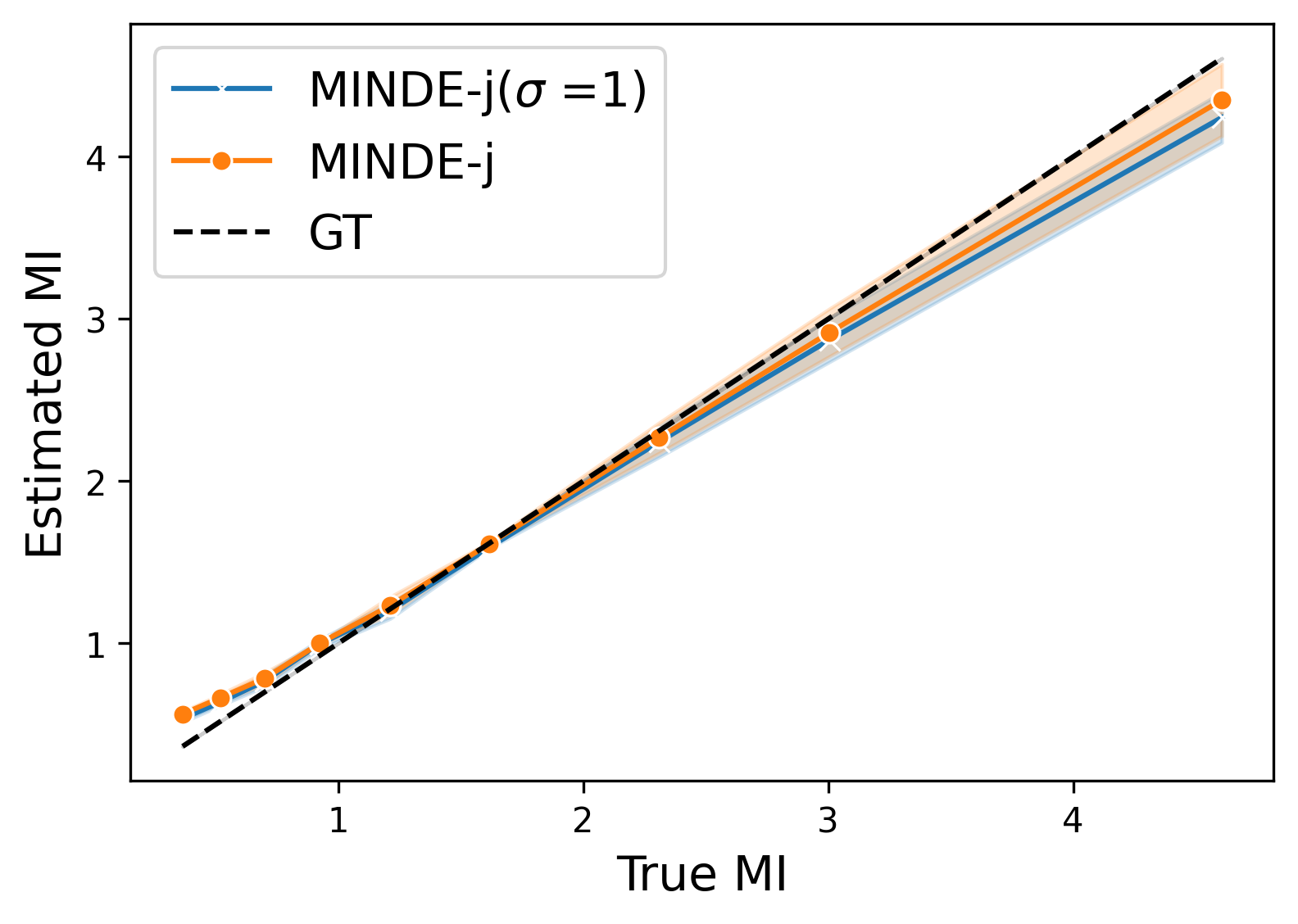}
\end{subfigure}
    \caption{ MI estimation results for \gls{MINDE}-j on 3 variables }
    \label{3vars-fig}
\end{figure}

In particular, we explore the case where the three random variables are distributed according to a multivariate Gaussian distribution: $A \sim  \gamma_1$, $B=A+N_1$ (with $N_1\sim \gamma_{\epsilon}$) and $C=A+N_2$ (with $N_2\sim \gamma_{\rho}$). By changing the values of the parameters, it is possible to change the value of the interaction information. We report in \Cref{3vars-fig} the estimated values versus the corresponding ground truths, showing that \gls{MINDE} variants can be effectively adapted for the task of information estimation between more than two random variables.




\end{document}